\documentclass[10pt,twocolumn,letterpaper]{article}

\usepackage[pagenumbers]{wacv} 

\usepackage{graphicx}
\usepackage{amsmath}
\usepackage{amssymb}
\usepackage{booktabs}
\usepackage{adjustbox}
\usepackage{multirow}
\usepackage{lipsum}

\usepackage[pagebackref,breaklinks,colorlinks]{hyperref}

\usepackage[capitalize]{cleveref}
\crefname{section}{Sec.}{Secs.}
\Crefname{section}{Section}{Sections}
\Crefname{table}{Table}{Tables}
\crefname{table}{Tab.}{Tabs.}
\newcommand{\revision}[1]{{\color{black}#1}}

\begin{document}

\title{AgroGPT\includegraphics[width=0.04\linewidth]{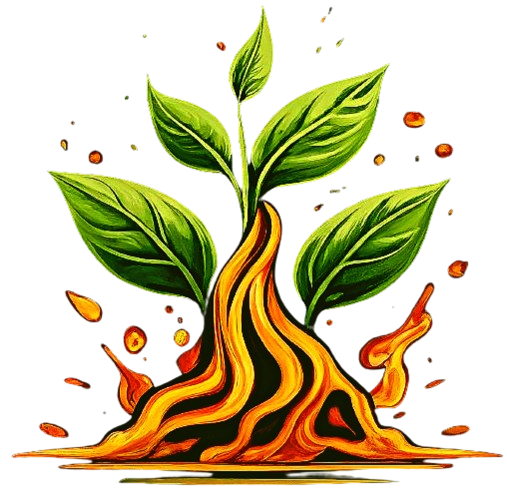}: Efficient Agricultural Vision-Language Model with Expert Tuning}

\author{%
  Muhammad Awais$^{1,2}$, Ali Husain Salem Abdulla Alharthi$^{1}$, Amandeep Kumar$^{1, 3}$, \\ Hisham Cholakkal$^1$, Rao Muhammad Anwer$^{1}$ \\
  1 Mohamed bin Zayed University of Artificial Intelligence, Abu Dhabi, UAE \\
  2 Georgia Institute of Technology, Atlanta, GA, USA  \\
  3 Johns Hopkins University, Baltimore, USA \\
   \texttt{aawais@gatech.edu}\\ \texttt{\{ali.alharthi, amandeep.kumar, hisham.cholakkal, rao.anwer\}@mbzuai.ac.ae}
}


\maketitle

\begin{abstract}
    {Significant progress has been made in advancing large multimodal conversational models (LMMs), capitalizing on vast repositories of image-text data available online. Despite this progress, these models often encounter substantial domain gaps, hindering their ability to engage in complex conversations across new domains. Recent efforts have aimed to mitigate this issue, albeit relying on domain-specific image-text data to curate instruction-tuning data. However, many domains, such as agriculture, lack such vision-language data. In this work, we propose an approach to construct instruction-tuning data that harnesses vision-only data for the agriculture domain. We utilize diverse agricultural datasets spanning multiple domains, curate class-specific information, and employ large language models (LLMs) to construct an expert-tuning set, resulting in a 70k expert-tuning dataset called AgroInstruct. Subsequently, we expert-tuned and created AgroGPT, an efficient LMM that can hold complex agriculture-related conversations and provide useful insights. We also develop AgroEvals for evaluation and compare {AgroGPT's} performance with large open and closed-source models. {AgroGPT} excels at identifying fine-grained agricultural concepts, can act as an agriculture expert, and provides helpful information for multimodal agriculture questions. The code, datasets, and models are available at \url{https://github.com/awaisrauf/agroGPT}.}
\end{abstract}

\section{Introduction}
\label{sec:intro}

\begin{figure}[t]
    \centering
    \includegraphics[width=0.5\textwidth]{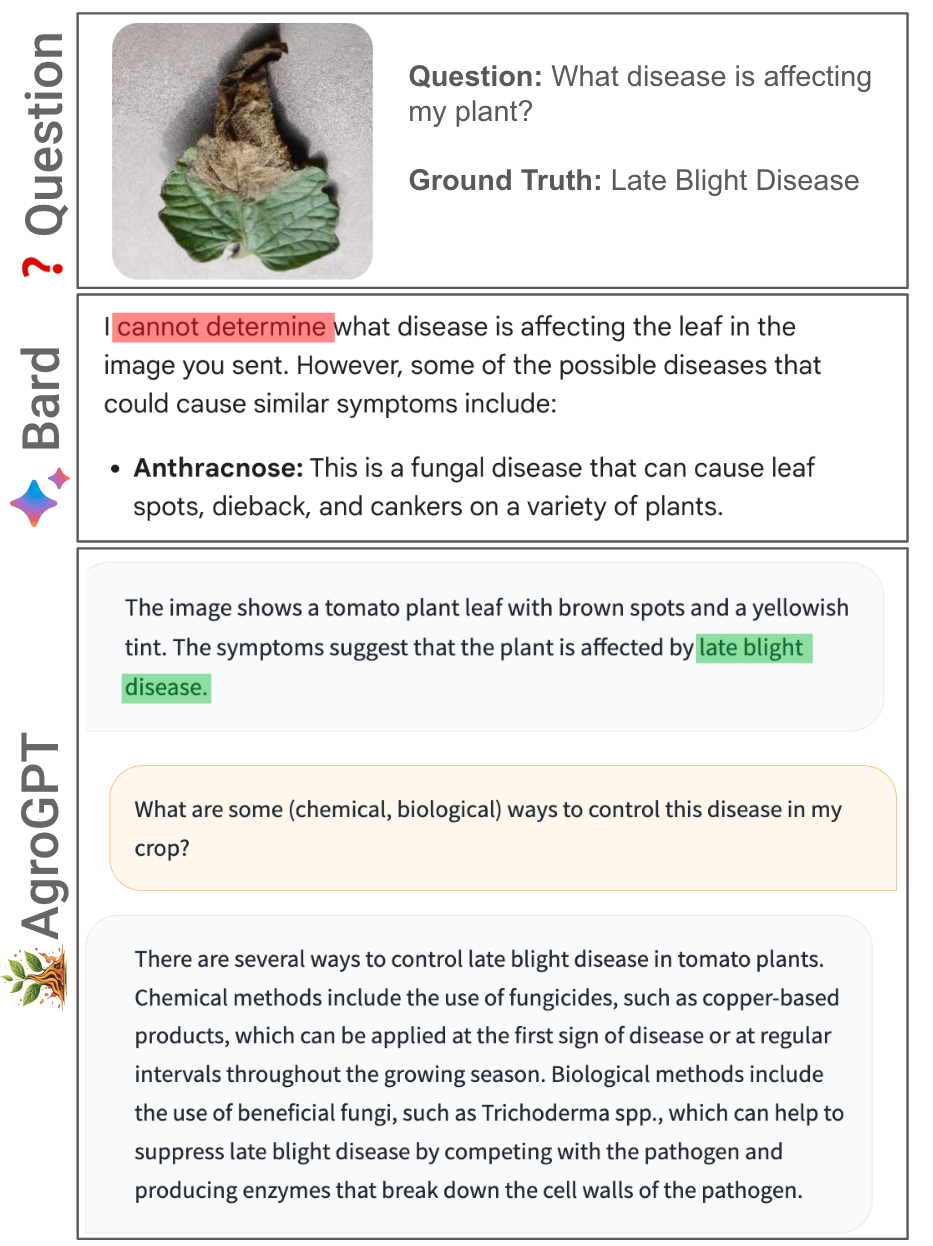}
    \caption{Comparison of AgroGPT with Bard to identify and provide information about a disease from leaf image. Our model can correctly identify the disease and the effect of the disease on the plant.}
    \label{fig:demo_comparison}
\end{figure}

The widespread availability of joint image-text data across the internet has catalyzed significant advancements in large multimodal models (LMMs), such as CLIP~\cite{radford2021learning}, GPT-4~\cite{achiam2023gpt}, Bard~\cite{nyberg2022bard}, and LLaVA~\cite{liu2023visual}. These models, when instruction-tuned, have demonstrated remarkable zero-shot performance and multi-turn conversational capabilities~\cite{ouyang2022training}. However, due to the inherent domain gap, their efficacy often falters in domains requiring specialized expertise, such as biomedicine, climate science, and geosensing, resulting in incorrect responses and hallucinations. To address this limitation, several fields have successfully leveraged existing domain-specific vision-language datasets to create domain-specific instruction-tuning data, enhancing model performance in specialized areas~\cite{mullappilly-etal-2023-arabic, kuckreja2023geochat, Omkar2023XrayGPT}. For instance, BioMedLLaVA utilized large vision-language data PMC-15M~\cite{zhang2023biomedclip} to curate biomedical instruction-tuning data resulting in LLaVA-Med~\cite{li2023llavamed}.

Despite these advancements, many domains, including agriculture, possess vision-only datasets but lack high-quality joint image-text data suitable for creating domain-specific instruction-tuning data.
To address this challenge, we introduce AgroGPT, the first approach to extending multimodal instruction tuning to the agricultural domain without requiring joint image-text data. Our method consists of three main components: a pipeline to generate expert-tuning data, an efficient multimodal model trained on AgroInstruct to bridge the domain gap, and an evaluation framework for domain-specific evaluation of the expert model. 
First, we propose a pipeline for synthesizing expert-level instruction-tuning data from vision-only datasets, effectively bridging the multimodal data gap in agriculture. This resulted in \emph{AgroInstruct}, a 70k expert-tuning dataset for agriculture. 
Second, \emph{AgroGPT}, we present an efficient, open-source expert agricultural conversational model, trained on the AgroInstruct to bridge the domain gap. Our proposed model is capable of holding complex conversations about the input agricultural images and can provide useful insights. Finally, we develop \emph{AgroEvals}, a simple visual question-answering framework to evaluate model performance in agricultural contexts.

Our main contribution is  AgroInstruct, a pipeline for curating expert-level instruction-tuning data solely from vision-only datasets specifically designed to enhance LMMs in the agricultural domain. To this end, we aggregate six diverse image-only datasets spanning four agricultural domains including plant diseases, weeds, farm insects, and fruits. 
Next, to transform the vision-only data into rich expert-tuning examples, we synthesize context-grounded image descriptions using state-of-the-art general-purpose LMMs, extract relevant image attributes, and collect class-specific background knowledge from agricultural sources.
Finally, we develop domain-specific instruction-following examples and system prompts tailored to agricultural contexts. These elements are fed into language-only LLMs to generate diverse and contextually relevant conversations.
Further, we implement a rule-based system leveraging dataset attributes to generate concise question-answer pairs, enhancing the variety and specificity of the instruction-tuning data.
Through this pipeline, we bootstrapped AgroInstruct, an extensive and diverse multimodal instruction-tuning dataset comprising 70k high-quality conversations.

Furthermore, we extend the two-step visual instruction tuning~\cite{liu2023visual} to include expert tuning and train efficient, conversational models (AgroGPT-3B and AgroGPT-7B) on AgroInstruct. We develop six visual question-answering (VQA) tasks (AgroEvals) across three groups to evaluate the model's capabilities.
To demonstrate the effectiveness of our method, we present comprehensive quantitative and qualitative results, comparing our models with several open and closed-source models, such as ChatGPT, Bard, and LLaVA-34B. On VQA tasks, our model significantly outperforms the baselines and ChatGPT in identifying fine-grained agricultural concepts. Additionally, AgroGPT excels at multi-turn, complex, domain-specific conversations, providing better guidance than several state-of-the-art general models. A demonstration of our model's conversation ability compared with Bard is shown in Figure~\ref{fig:demo_comparison}.

\begin{table}
\centering 
    \includegraphics[width=0.5\textwidth]{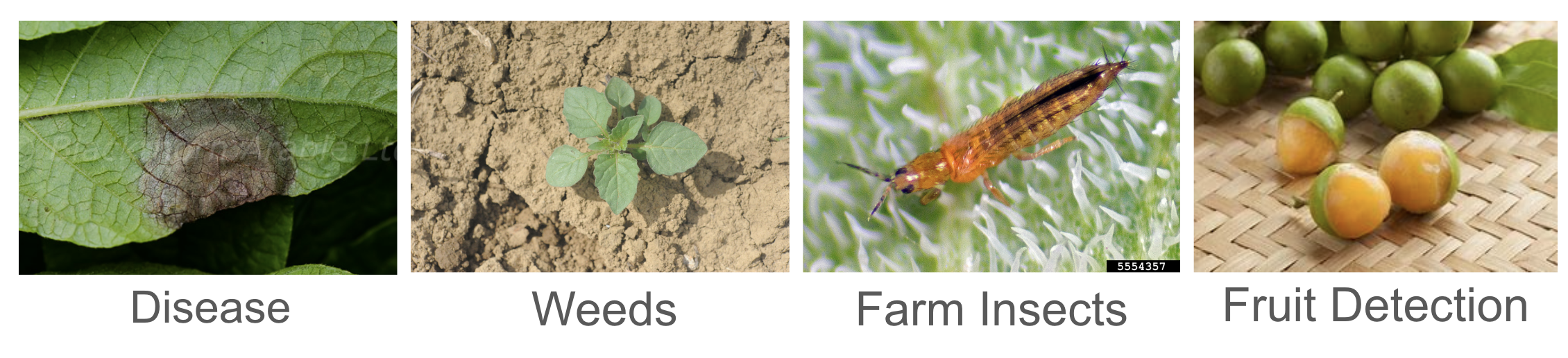}
    \centering
    \resizebox{0.5\textwidth}{!}{
    \begin{tabular}{lrr}
    \toprule
       Dataset & Num of Imgs & Classes   \\
    \toprule
        Cotton Leaf Disease~\cite{D3v} & 4800 & 6 \\
        PlantVillage~\cite{hughes2015open}  & 54303 & 38  \\
        PlantDoc~\cite{10.1145/3371158.3371196}  & 2598 & 30 \\
        \midrule
        Fruits100~\cite{D4v}  & 40000 & 100 \\
        \midrule
        Early Weed Detection~\cite{di2017automatic} & 1000 & 4 \\
        \midrule
        Farm Insects~\cite{D5v} & ~6000 & 12\\
        \bottomrule
    \end{tabular}}
    \caption{Four agricultural domains (disease, weeds, insects, and fruits) and statistics of vision-only datasets used to create our expert-tuning AgroInstruct data.}
    \label{tab:agri_datasets}
\end{table}

\begin{figure*}
    \centering
    \includegraphics[width=1\textwidth]{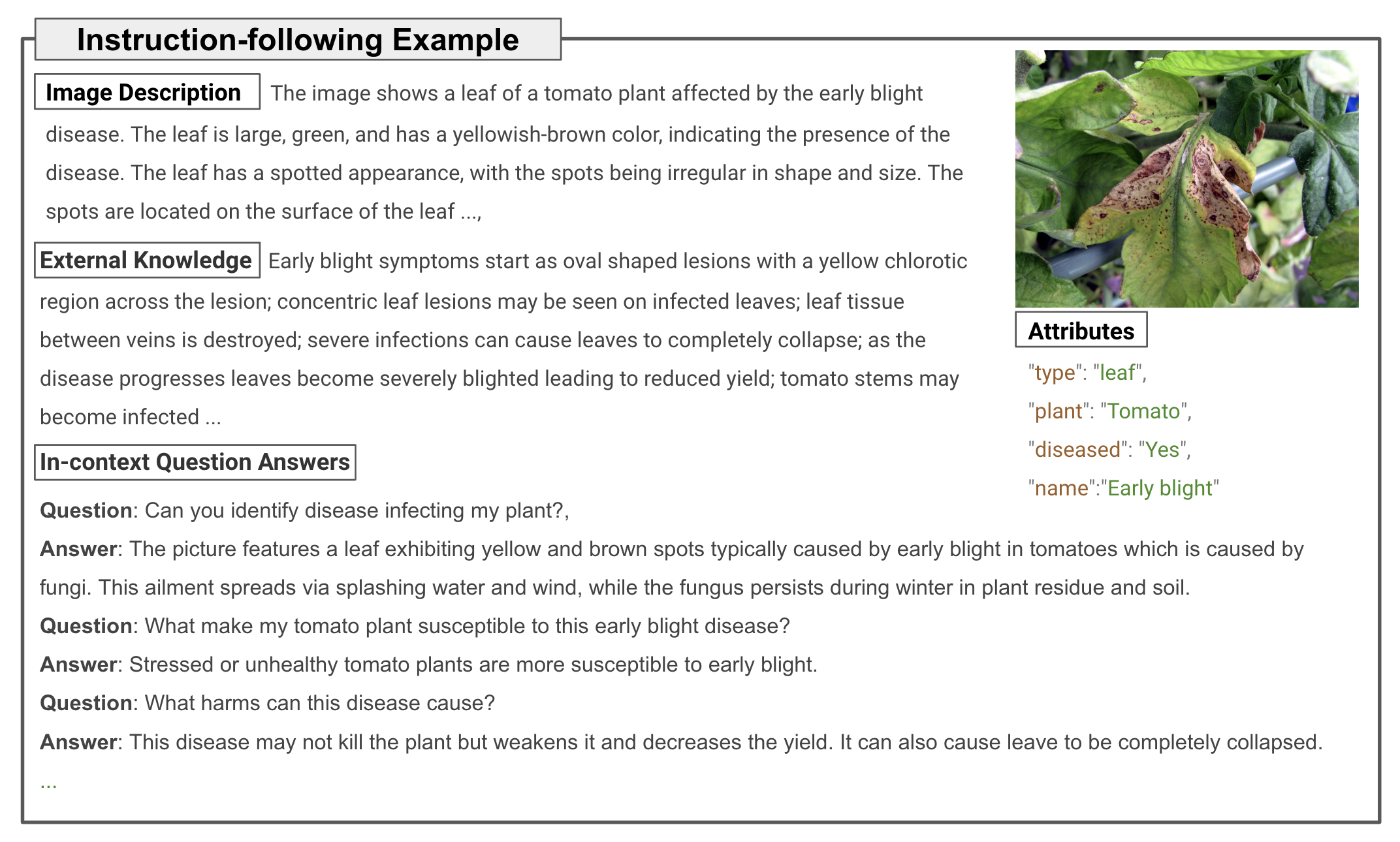}
    \caption{An illustration of instruction-following examples used to generate expert-tuning from vision-only agriculture data.}
    \label{fig:conversation_context}
\end{figure*}

\begin{figure}
    \centering
    \includegraphics[width=0.5\textwidth]{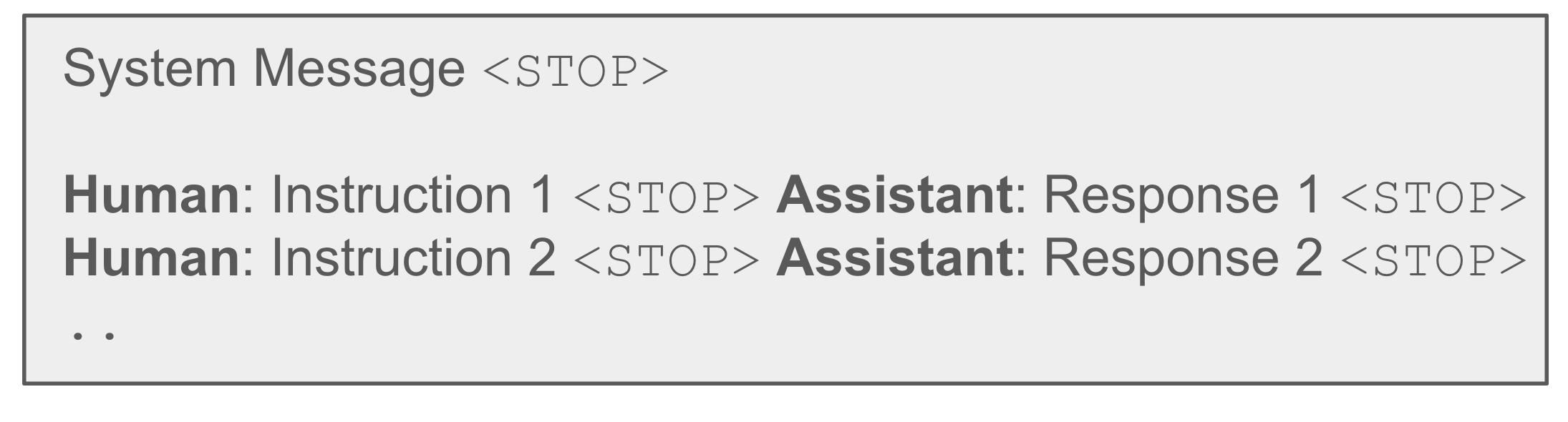}
    \caption{An illustration of sequencing of conversation before being fed to LLM, following LLaVA and LLaVA-Med~\cite{liu2023visual, li2023llavamed}.}
    \label{fig:prompt_format}
\end{figure}

\section{Related Work}
\noindent\textbf{General Large Language Models (LLMs) and Large Multimodal Models (LMMs). }
Large Language Models (LLMs) possess remarkable capabilities to understand and generate text that resembles human language. This enables them to tackle various tasks effectively, with GPT-4 \cite{peng2023instruction}, Llama-2 \cite{touvron2023llama}, and Gemini \cite{team2023gemini}. Recently, these LLMs are extended to multimodal inputs~\cite{awais2023foundational} by aligning visual information in the LLM embedding space for which a pre-trained visual backbone \cite{dosovitskiy2020image} is used to process visual data, a large language model \cite{chiang2023vicuna} for understanding user instructions and generating responses, and a cross-modal connection MLP \cite{liu2023improved} to integrate visual information with language models. The impressive improvements in tasks related to language instruction following and visual reasoning in natural scenes are recently shown in LLaVA \cite{liu2023visual}, MiniGPT-4 \cite{zhu2023minigpt} \cite{chen2023minigptv2}, InstructBLIP \cite{liu2023improved}, Otter \cite{li2023otter}.

\noindent\textbf{Expert Foundational Models. }
Due to their training on a large corpus of data, recent foundational models have achieved remarkable few-shot and zero-shot learning capabilities \cite{brown2020language, kojima2022large, wei2021finetuned}, leading to having impressive performance even in domains where the model is not specifically trained. However, these large models are not particularly designed for specific domain expertise.\revision{ Therefore, recently, a large number of effort has been made on open-sourced instruction-tuned LLMs to train a domain-specific models like Mini-ClimateGPT \cite{mullappilly-etal-2023-arabic}, GeoChat \cite{kuckreja2023geochat, hu2023rsgpt}, XrayGPT \cite{Omkar2023XrayGPT}, ClimateGPT \cite{thulke2024climategpt} and VideoGPT \cite{Maaz2023VideoChatGPT, rasheed2023glamm}. However, unlike these models, we introduce a pipeline to construct expert-tuning data from existing image-only datasets and develop efficient AgroGPT (3 and 7B), which can hold complex conversations about fine-grained agricultural concepts. Recently, some works have also explored CLIP-like contrastive models for agriculture for better pre-training for downstream agricultural tasks~\cite{nawaz2024agriclip}.}

\revision{
\noindent\textbf{Deep learning in Agriculture. }
Supervised learning-based approaches have mainly been explored for agriculture, such as vision-based datasets for different important tasks in agriculture (plant diseases~\cite{hughes2015open,10.1145/3371158.3371196, plant_disease_recognition_2021, sugarcane_leaf_disease_2024, potato_leaf_disease_2020}, harmful insects~\cite{mohan2022insects}, fruits~\cite{olimov2024fruits}, and weeds identification~\cite{di2017automatic}. However, the agriculture field lacks any comprehensive image-text datasets. Similarly, deep learning models have also been proposed to solve these tasks~\cite{atila2021plant,
feng2022vegetable,
thakur2023vgg}. Recently, PLLaVA~\cite{yang2024pllama} was introduced for plants. However, it is limited to text-based questions. This paper proposes an AgroInstruct pipeline to convert vision-only datasets into expert-tuning datasets and train a large model capable of identifying fine-grained agricultural concepts and holding complex conversations. 
}

\section{Agricultural Expert-tuning}
In the following sections, we first explain our efficient pipeline for curating expert tuning data to instill agricultural knowledge in Large Multimodal Models (LMMs). We utilize classification-only image datasets and class-specific information to synthesize the data. Next, we describe the architecture of our LMMs, followed by the training process using the expert tuning dataset. This tuning is performed after the vision-language alignment and general visual instruction tuning stages~\cite{liu2023visual}.

\subsection{Curation of AgroInstruct for Expert Tuning}

The agricultural domain has several vision-only datasets (as depicted in Figure~\ref{fig:conversation_context}), but it lacks multimodal data, presenting a challenge for developing instruction-tuning datasets. To address this issue, we propose a solution that leverages vision-only datasets from four different domains (diseases, weeds, insects, fruits) as shown in Tab.~\ref{tab:agri_datasets} to synthesize expert tuning data called AgroInstruct following visual instruction tuning~\cite{liu2023visual} and LLaVA-med~\cite{li2023llavamed}.

\begin{figure}
    \centering
    \includegraphics[width=0.5\textwidth ]{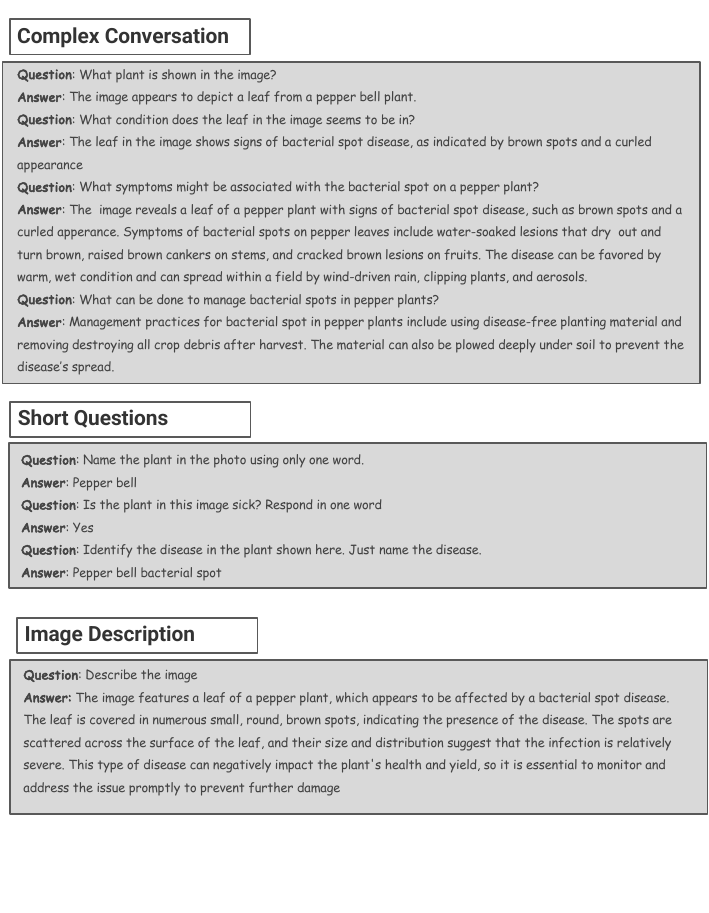}
    \vspace{-50pt}
    \caption{An example of three types of conversations in the AgroInstruct dataset: complex, multi-turn, short questions, and image descriptions. Zoom in for a better view.}
    \label{fig:sample_dataset}
\end{figure}

AgroInstruct consists of three types of questions: complex multi-turn conversations, simple questions, and image descriptions. First, we generate an image description with generic LMMs by providing it with the context of the dataset and image-specific attributes. Next, to generate multi-turn complex conversations, we utilize an open-source LLM to generate expert-tuning data by providing it with context-conditioned descriptions, class-specific external background information from reputable agricultural resources, and in-context examples of conversations as shown in Figure~\ref{fig:conversation_context}. This is provided as input into a Large Language Model (LLM) alongside appropriate system prompts. Finally, we utilize attributes of image-only datasets to generate simple questions, such as identifying plants and diseases.  Our expert-tuning data, AgroInstruct, consists of 70k conversations belonging to three types (10K descriptions, 35 complex questions, and 35k simple questions). An example of these three types of questions is provided in Figure~\ref{fig:sample_dataset}. 

In the following, we elaborate on each step. 

\begin{figure}
    \centering
    \includegraphics[width=0.5\textwidth]{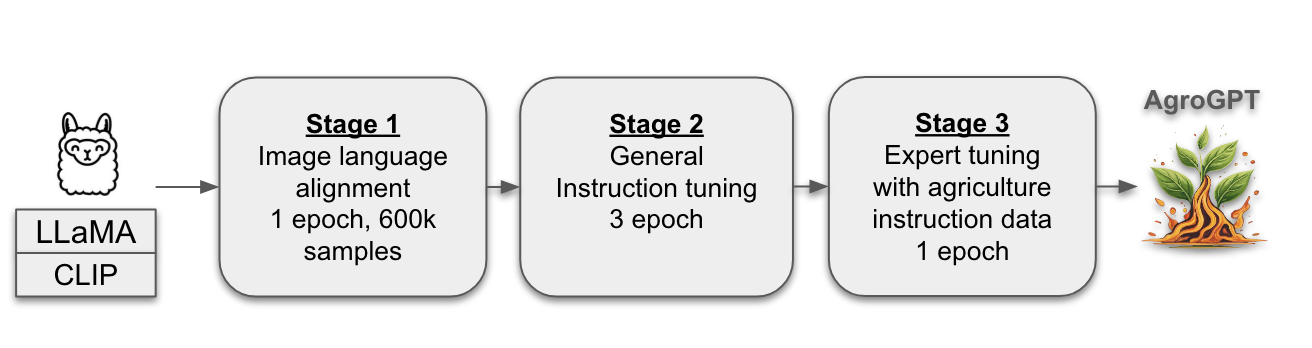}
    \caption{An overview of AgroGPT three-stage training process to convert an image encoder and language-only LLM to a domain expert. The first two stages consist of vision-language concept alignment and general instruction-tuning, which is based on LLaVA~\cite{liu2023visual}.}
    \label{fig:training}
\end{figure}

\noindent\textbf{Context-grounded Image Descriptions. }
First, we generate context-grounded image descriptions by prompting a vision-language model to describe the input based on dataset-specific attributes such as plant name, disease name, and whether it's a plant or a leaf. We construct a system of prompts to instruct models to utilize context and generate specific and field-specific information. We devised 10 questions to describe the image following MedLLaVA~\cite{li2023llavamed}. This process resulted in the construction of approximately 35K descriptions spanning four domains. We only utilize 10k descriptions in our AgroInstruct data, while the rest are utilized as context for the generation of complex conversations explained next.

\begin{table}
    \centering
    \begin{minipage}{0.48\textwidth}
    \resizebox{1\textwidth}{!}{
    \begin{tabular}{p{0.18\textwidth}p{0.82\textwidth}}
    \toprule
         & \includegraphics[width=0.25\textwidth]{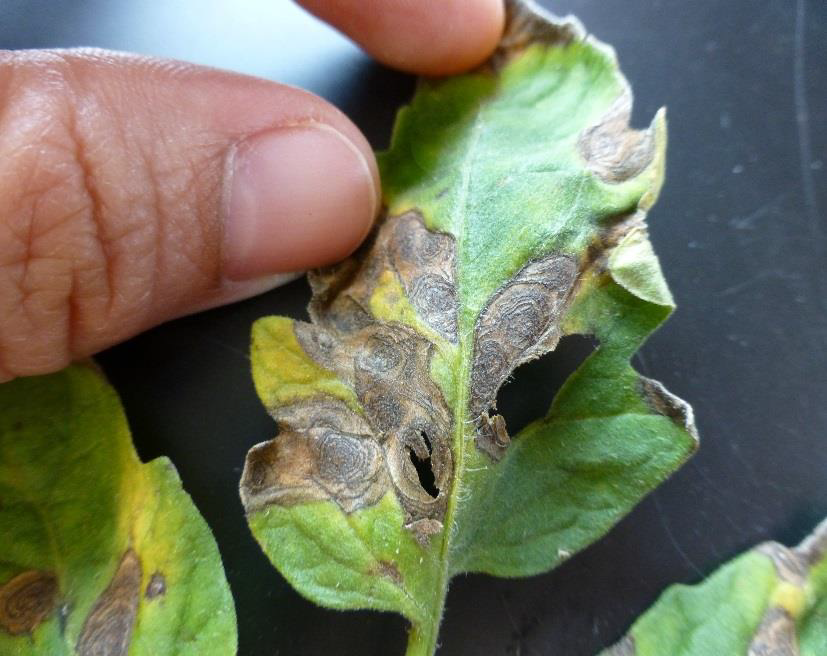}   \\
         \toprule
         \textbf{User}:     & What disease is affecting my plant? \\
         \textbf{AgroGPT}:  & The plant in the image is affected by {\color{blue}early blight}, a fungal disease. \\
         \textbf{User}:     & How can I control this disease? \\
         \textbf{AgroGPT}:  & The disease can be controlled by applying appropriate fungicides at the first sign of disease, removing and destroying infected plant debris, and practicing crop rotation. It is also important to maintain good plant hygiene and avoid over-application of nitrogen fertilizer. \\
    \midrule
         \textbf{Bard}:      & Based on the image you sent, the plant is most likely affected by a bacterial disease. There are a {\color{red}few possibilities}, including: Bacterial leaf spot, Xanthomonas spot, and Erwinia blight. \\
    \midrule
         \textbf{LLaVA-34B}:  & The plant in the image appears to be affected by a disease that causes leaf spots and necrotic tissue. The spots are brown and have a concentric ring pattern, which is characteristic of {\color{red} a fungal disease}. \\
    \bottomrule
    \end{tabular}}
    \end{minipage}
    \caption{Comparison of our efficient AgroGPT-3B's abilities with Bard, ChatGPT4o, and LLaVA-34B on open-ended questions for an image taken from web. AgroGPT-3B correctly identifies the disease (early blight) and suggests control methods, while Bard and ChatGPT-4o decline to answer, and LLaVA-34B provides an incorrect response. }
    \label{tab:quant_outside}
    \end{table}

\noindent\textbf{Complex, Multi-turn, and Open-ended Conversations. } 
We need a dataset containing multi-turn conversations to equip our model with the capability to answer complex, open-ended, and agriculture-related questions about an image. To this end, we craft a diverse set of conversations featuring multi-round conversations using a language-only Large Language Model (LLM) named Mistral~\cite{jiang2023mistral}. To guide LLM in generating field-specific and to-the-point conversations, we utilize the following four as context for each input image. 

\begin{figure*}[t]
    \centering
    \includegraphics[width=1\linewidth]{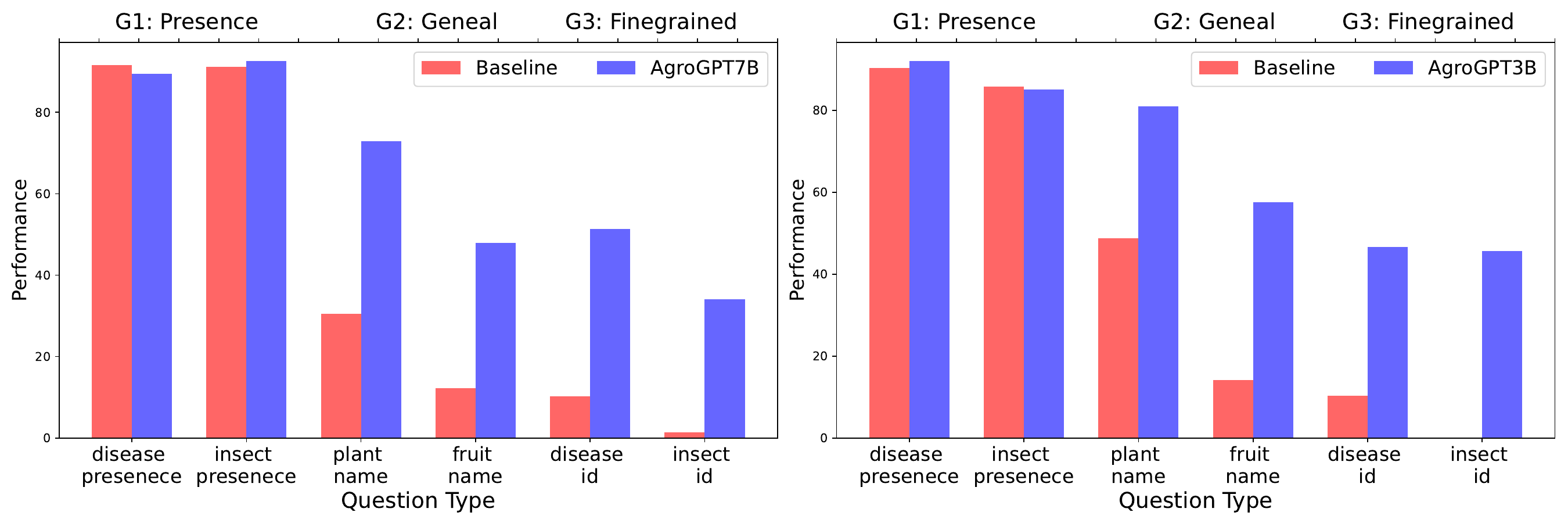}
    \caption{Comparison of our method's ability to improve agricultural knowledge of baseline on two versions of AgroGPT. General instruction-tuned models work well for general knowledge (e.g., is this leaf diseased?) but fail on fine-grained concepts such as disease names or insect names. Expert tuning significantly improves this. }
    \label{fig:main_results}
\end{figure*}

\begin{itemize}
    \item \noindent\textbf{Image Attributes. } Image-only datasets (Table~\ref{tab:agri_datasets}) have attributes for each image, such as plant name, disease name, fruit name, etc. These attributes are added in the context of LLM to facilitate its understanding of the nature of the input.

    \item \noindent\textbf{External Information. } To further enable the model to understand nuanced information about the topic of the input image, we curated class-specific information from reputable agricultural external resources for each class. Our dataset consists of 202 classes, and we curate, on average, a page-long external information for each class. We add external resources in the Appendix.

    \item \noindent\textbf{In-context Examples. } We created a few in-context examples for each domain to create a diverse set of multi-turn question-answers. These in-context examples are meant to guide LLM in the generation of question-and-answer pairs. Further, we also want the model to use the information provided in the attributes, descriptions, and external information. These in-context examples are designed in such a way that they correlate information with answers, hence guiding LLM to look into the provided information. 

    \item \noindent\textbf{System Prompt. } Finally, we generated a system prompt for each domain to instruct the model to generate comprehensive conversations for that domain. Overall, we follow MedLLaVA's~\cite{li2023llavamed} technique to create these prompts. 

\end{itemize}

We provide an example in Figure~\ref{fig:conversation_context}. We provide this context to an LLM (Mistral) and system prompt to create 3-5 questions per image. These conversations include questions to identify specific elements within the image and answer domain-specific questions about the identified object. We produced approximately 35k multi-round, complex, agriculture-related conversations using this method.

\noindent\textbf{Simple Question Answers}
In addition to the complex, multi-round conversations, we have augmented our dataset with rule-based simple question-answer pairs. These questions are designed to help the model accurately identify specific elements within the image and provide concise answers.
To create these questions, we leverage dataset attributes such as the plant name, health status (diseased or healthy), disease name, insect names, fruit names, etc. Subsequently, we generate several variations for each type of question, prompting the model to respond with a single word regarding the specified attribute. For example, for the disease attribute, one question is: "What disease does the plant in the image have? Provide the name of the disease only." We have generated 35k such question-answer pairs.

\subsection{Architecture of AgroGPT}

For our baseline model, we utilized two types of model from the LLaVA family, comprising a visual encoder, a projection layer, and an LLM. For our AgroGPT, we employed two different models: a 7B LLaVA model~\cite{liu2023visual} and an efficient 3B Mipha model~\cite{zhu2024mipha}. These models are chosen to suit different computational requirements and performance considerations for downstream applications. LLaVA and Mipha use pre-trained CLIP ViT-L / 14~\cite{radford2021learning} and SigLIP~\cite{zhai2023sigmoid} as visual encoders. In our experiments, we found SigLIP better than the traditional CLIP model, even when a smaller LLM is used. In conjunction with the projection layer, this visual encoder transforms an image into the language embedding space, producing visual tokens. These image tokens are combined with language instruction tokens and fed into the LLM. Subsequently, the LLM generates text conditioned on the language instructions and the image. 

\subsection{Expert Tuning AgroGPT with AgroInstruct}

Our main goal is to instruction-tune a vision-language chat model that is expert in the agriculture field. To achieve this, we use the first two stages of visual instruction tuning~\cite{liu2023visual} and introduce an expert tuning stage, as shown in Figure~\ref{fig:training}. Specifically, the first stage consists of visual-language concept alignment on general images, and the second stage consists of instruction tuning on general vision-language data. We introduce a third expert tuning stage to instill agriculture-related expertise in the model. We kept the visual encoder frozen for expert tuning and trained the projection layer and LLM weights on our AgroInstruct data. For training, we converted all three formats of instructions (descriptions, small questions, multi-turn, and complex questions) into the same conversation format. Then, we randomly sampled conversations, added system messages as shown in Appendix Figure~\ref{fig:prompt_format}, and trained LLM on its original auto-regressive objective. In the ablations, we also compare the effect of expert tuning after stage 1 and stage 2. Generally, expert tuning after stage 2 results in better results. 

\begin{table}
    \centering
    \begin{minipage}{0.48\textwidth}
    \resizebox{1\textwidth}{!}{
    \begin{tabular}{p{0.15\textwidth}p{0.8\textwidth}}
    \toprule
         &  \includegraphics[width=0.28\textwidth]{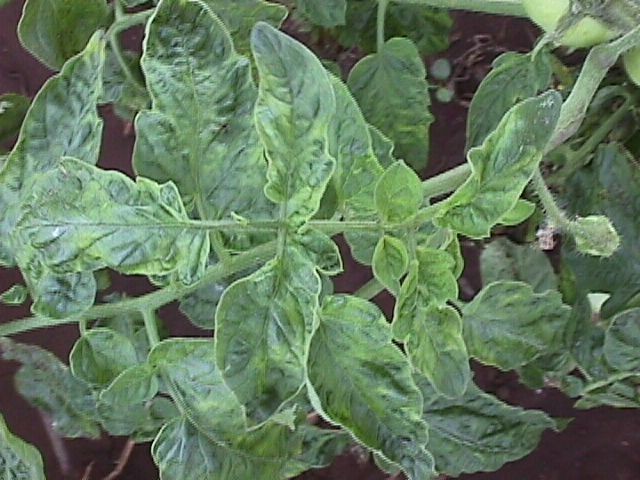}   \\
    \midrule
     \textbf{User}:     & What is the disease affecting the leaves? \\
      \textbf{AgroGPT}:   & The leaf in the image appears to be affected by {\color{blue}tomato leaf curl virus}.\\
     \textbf{User}:      & How can I control this disease? \\
     \textbf{AgroGPT}:  & Tomato Leaf Curl Virus is a viral disease that can be controlled planting resistant varieties, removing infected plants and burning them, avoiding the use of infected seedlings, and practicing crop rotation.\\

     \midrule
      \textbf{Bard}:    & Based on the image you sent, {\color{red}it is difficult to say} for sure what disease is affecting the leaf However, there are a few possibilities, including: Early blight(..) Septoria leaf spot(..), Bacterial leaf spot (..) It is also possible that the leaf is damaged by insects or other pests. \\
     \midrule
    \textbf{LLaVA-34B}: & The plant in the image appears to be affected by a 
               disease that is causing the leaves to become 
              discolored and possibly wilted. The symptoms 
              include yellowing and browning of the leaf edges,
              which  {\color{red}could indicate a variety of issues} such
              as nutrient deficiency, water stress, or a fungal or 
               bacterial infection \\
    \bottomrule
    \end{tabular}}
    \end{minipage}
    \caption{Comparison of efficient AgroGPT-3B with Bard and 34B LLaVA on open-ended questions for an image taken from the held-set of PlantDoc dataset~\cite{10.1145/3371158.3371196}. AgroGPT can correctly identify the disease and suggest several methods to control it in the crop. In contrast, Bard and LLaVA-34B are unable to identify the name. }
    \label{tab:quant_disease_test}
\end{table}
\begin{table}
    \centering
\resizebox{0.48\textwidth}{!}{
    \begin{tabular}{p{0.075\textwidth}p{0.4\textwidth}}
        \toprule
         & \includegraphics[width=0.15\textwidth]{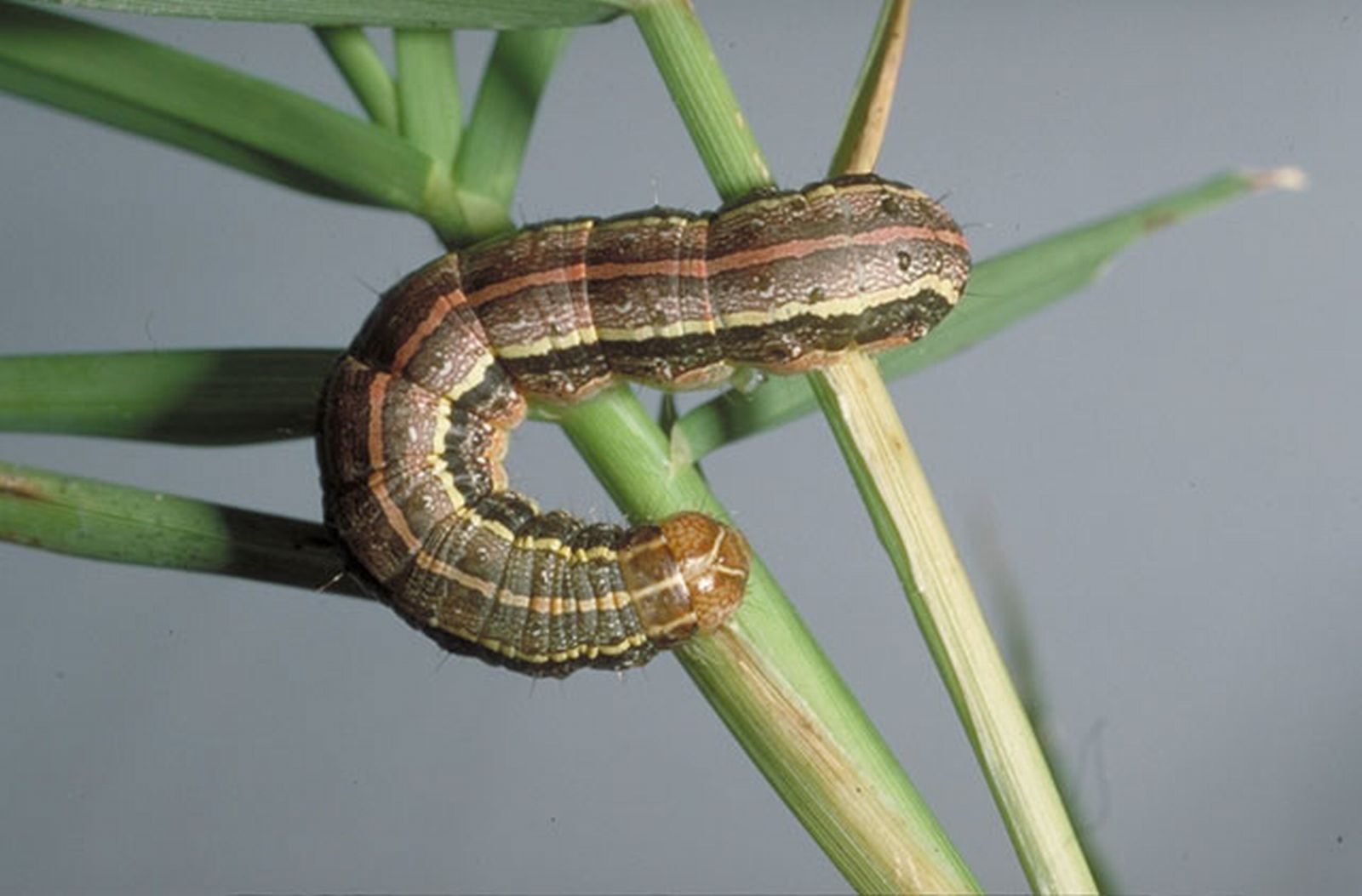}   \\
         \midrule
         \textbf{User}:    & What insect is in this image? \\
         \textbf{AgroGPT:} & The insect in the image is a {\color{blue}fall armyworm}.\\
         \textbf{User}:    & How can it affect my crop? \\
         \textbf{AgroGPT}: & Fall armyworms are a type of caterpillar that can cause significant damage to crops, particularly corn and soybeans. They feed on the leaves, stems, and fruits of these plants, which can lead to reduced yield and quality. In severe cases, the infestation can cause complete crop failure. The damage caused by fall armyworms can be difficult to control and can result in significant economic losses for farmers.\\

         \midrule
          \textbf{Bard}:    &The insect in the image you sent appears to be a caterpillar, but it is {\color{red}difficult  to identify} the exact 
 species without more However, based on  the visible characteristics, here are some possibilities:  Mythimna separata(..), Fall armyworm (..), Prodenia spp(..), Texas armyworm caterpillar(..) .. \\
         \midrule
        \textbf{LLaVA34B}: & The insect in the image is a {\color{red}caterpillar}.\\
         \bottomrule
    \end{tabular}}
        \caption{Comparison of efficient AgroGPT-3B with Bard and 34B LLaVA on open-ended questions for an image taken from the holdout-set of Farm Insects dataset~\cite{D5v}. AgroGPT can correctly identify the name of the insect and asses the potential harm it can do. In contrast, Bard is uncertain and provides several possibilities, and LLaVA34B incorrectly identifies it as a caterpillar. }
    \label{tab:quant_insect_test}
\end{table}

\section{Experimental Results}

\noindent \textbf{Model Training. }We use Phi-2-2.7B~\cite{javaheripi2023phi}
 and Vicuna-7B~\cite{zheng2023judging} as our LLMs and CLIP visual encoder ViT-L/14~\cite{radford2021learning} and SigLIP~\cite{zhai2023sigmoid} with resolution 336 as our image encoders. We used Mipha~\cite{zhu2024mipha} and LLaVA1.5~\cite{liu2023visual} provided pre-trained weights, which have been trained with visual instruction tuning. Unless stated otherwise, we applied our expert tuning after image-text alignment (stage 1) and visual instruction tuning (stage 2).  We use AdamW ~\cite{loshchilov2017decoupled} to optimize along with the cosine learning rate scheduler to train our model. Unless otherwise stated, we train our model for 1 epoch with a batch size of 8 or 16, depending on computational resources. For our training, we utilize 4 A6000 GPUs.  

\noindent\textbf{Expert Tuning Data. }To generate our expert tuning, we used two different models. First, we use Mistral7B~\cite{jiang2023mistral} to generate instruction tuning, following the method proposed in the previous section. Second, we use LLaVA-13B to create descriptions that are used as in-context information for Mistral7B, which is also explained in the previous section.

\subsection{Qualitative Results} 

To demonstrate the effectiveness of AgroGPT in identifying agricultural issues (diseases, insects, weeds) and guiding users, we hold conversations on several examples. First, we compared AgroGPT's performance with LLaVA-34B and Bard for crop diseases, farm insects, and weeds on held-out images. The results are shown in Tables~\ref{tab:quant_disease_test}, \ref{tab:quant_insect_test}. These conversations highlight AgroGPT's capability to identify fine-grained attributes (disease, insect, and weed names) and answer field-specific questions. For instance, AgroGPT accurately identifies diseases affecting the leaf, weed name, and farm insect in the images, provides suggestions for managing these issues, and explains its effect on the crop. In contrast, both Bard and LLaVA-34B either fail to identify fine-grained concepts or make incorrect identifications. 

To evaluate the generalizability of AgroGPT, we further collected several images from online agriculture forums where users have asked questions. We compare AgroGPT's capabilities to answer questions with LLaVA-34B, Bard, and ChatGPT4o. The conversations are presented in Table~\ref{tab:quant_outside}, and more results are relegated to the Appendix. AgroGPT can correctly identify diseases, assess potential damages, and suggest possible remedies. In contrast, both Bard and ChatGPT4o cannot answer it and LLaVA-34B hallucinates. For more details on quantitative evaluations, please refer to the Appendix. 

\subsection{Quantiative Evaluations} For quantitative evaluations, we create \emph{AgroEvals} by utilizing the test and a holdout set of datasets utilized in the generation of AgroInstruct (see Table~\ref{tab:agri_datasets}). Based on these datasets, we create six visual question-answering sets consisting of three groups that test the model's capabilities at different levels.
\textbf{Group 1} consists of questions to ask for the presence of a disease or insect in an image. Since these two attributes are general and coarse, a model trained on general instruction tuning (e.g., visual tuning) can easily answer these questions without expert tuning. \textbf{Group 2} consists of questions to detect general categories such as fruits or plant names. These two attributes are finer-grained than group 1 but are present in general instruction-tuning data. However, \textbf{Group 3} consists of questions that probe the model for fine-grained and field-specific questions, such as the name of the disease or insect. These concepts are less likely to be present in the general instruction-tuning data and hence general models struggle to answer these questions.
Further, for our quantitative evaluations, we use two versions of our models: AgroGPT7B, which is based on LLaVA1.5~\cite{liu2023visual} and AgroGPT3B, which is based on Mipha3B~\cite{zhu2024mipha}.

\begin{table}
    \centering
    \begin{adjustbox}{width=0.45\textwidth}
        \begin{tabular}{llccc}
        \toprule
        &\textbf{Category} & \textbf{Baseline} & \textbf{Stage2} & \textbf{Stage3} \\
        \midrule
        \multirow{ 2}{*}{Group1}&Disease Presence & 91.52 & 90.25 & 89.41 \\
        &Insect Presence  &91.16 & 96.59 & 92.51  \\
        \midrule
        \multirow{ 2}{*}{Group2}&Plant Name & 30.51 & 72.46 & 72.88 \\
        &Fruits Name & 12.30 & 47.38 & 47.96 \\
        \midrule
        \multirow{ 2}{*}{Group3}
        &Disease Id & 10.27 & 46.57 & 51.37 \\
        &Insect Id & 0.25 & 35.57 & 35.89\\
        \bottomrule
        \end{tabular}
    \end{adjustbox}
    \caption{Comparison of performing expert tuning after stage 1 (image-text alignment) vs. stage 2 (general instruction-tuning) for AgroGPT-7B.}
    \label{table:stage2vsstage3}
\end{table}

\noindent \textbf{Main Results}. In Figure~\ref{fig:main_results}, we compare the ability of our method to add field-specific knowledge with the baseline of visual instruction tuning across six types of questions from three groups. Models fine-tuned with our method perform on par with visual instruction tuning for Group 1 questions, which tests the model's coarse recognition abilities: presence and absence of disease and insects in a given image. However, our method performs significantly better for Group 2, which tests the model's more fine-grained recognition abilities (e.g., name of fruit or plant name.

However, baseline models (fine-tuned with visual instruction tuning) completely fail on fine-grained questions from Group 3, which tests more fine-grained and relevant concepts such as identifying diseases or insects. On the other hand, our proposed expert tuning can instill these fine-grained concepts in the model and significantly improve their performance. For instance, both 3B and 7B baseline Mipha and LLaVA models completely fail to identify insects, while our method significantly improves this performance. Interestingly, AgroGPT3B performs better on fine-grained questions compared to AgroGPT7B. We hypothesize that this is due to the Mipha~\cite{zhu2024mipha} baseline using a more capable image encoder (SigLIP~\cite{zhai2023sigmoid}) compared to LLaVA's CLIP.

\revision{
\noindent \textbf{Comparison with ChatGPT. } We also quantitatively compare AgroGPT with ChatGPT for benchmark from Group 3 of AgroEvals. We chose Group 3 as it represents fine-grained agriculture knowledge (plant disease and farm insect identification). As shown in Table~\ref{tab:chatGPT}, AgroGPT outperforms ChatGPT by significant margins, such as on diseases, our model has a performance of 51.37\% compared with ChatGPT's 30.82\%. It is also important to note that the efficient baseline model (Mipha3B) shows very poor performance as it lacks expert knowledge about fine-grained agricultural concepts. 

\begin{table}[]
    \centering
    \begin{tabular}{lc|cc}
    \toprule
          &Type& Plant Diseases & Farm Insects  \\
    \midrule
    Baseline &ZeroShot  & 10.27 & 0.00 \\
    \textbf{ChatGPT}& Zeroshot & 30.82 & 32.65   \\
     AgroGPT &Zeroshot   & 51.37 & 45.58  \\
    \bottomrule
    \end{tabular} 
    \caption{\revision{Comparison of AgroGPT with baseline and ChatGPT. AgroGPT outperforms ChatGPT in terms of fine-grained agricultural knowledge, demonstrating superior performance in fine-grained knowledge of agriculture (Group 3).}}
    \label{tab:chatGPT}
\end{table}

\begin{figure}
    \centering
    \includegraphics[width=1\linewidth]{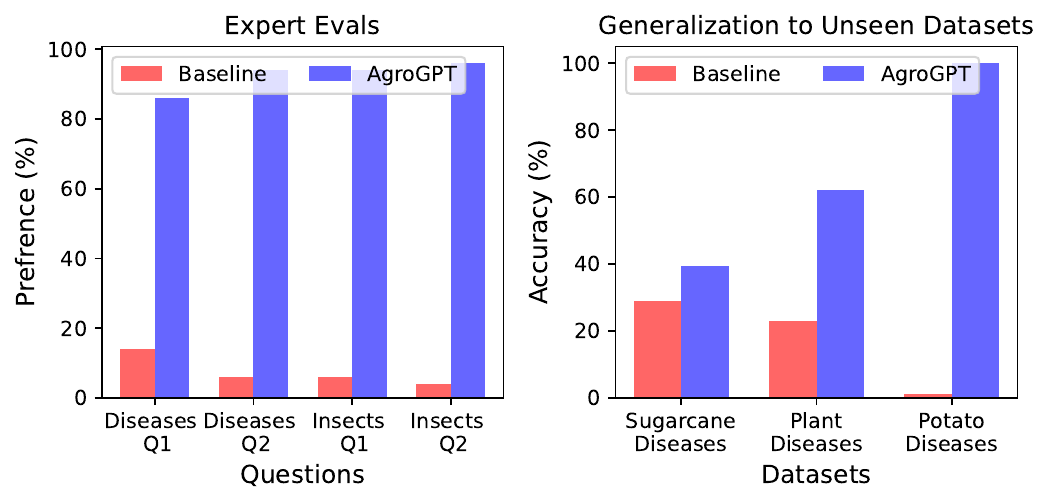}
    \caption{\revision{(a) Percentage of responses preferred by an expert for AgroGPT compared to the baseline. (b) Generalization performance of AgroGPT on datasets outside of its training set.}}
    \label{fig:post_rebuttal}
\end{figure}

\noindent \textbf{Performance on Unseen Datasets. } 
To evaluate our model's generalization to unseen agricultural concepts, we select three datasets: the Sugarcane Leaf Disease dataset~\cite{sugarcane_leaf_disease_2024}, the Potato Leaf Disease dataset~\cite{potato_leaf_disease_2020}, and the Plant Disease Recognition dataset~\cite{plant_disease_recognition_2021}. We then assessed AgroGPT and our baseline using the AgroEvals protocol described previously. As illustrated in Figure~\ref{fig:post_rebuttal}(b), AgroGPT demonstrates significantly higher performance, indicating its ability to generalize beyond the datasets used in expert tuning. However, it is crucial to note that AgroGPT may not perform as well on concepts that differ considerably, such as livestock, which are not part of AgroInstruct.

\begin{figure}
    \centering
\includegraphics[width=1\linewidth]{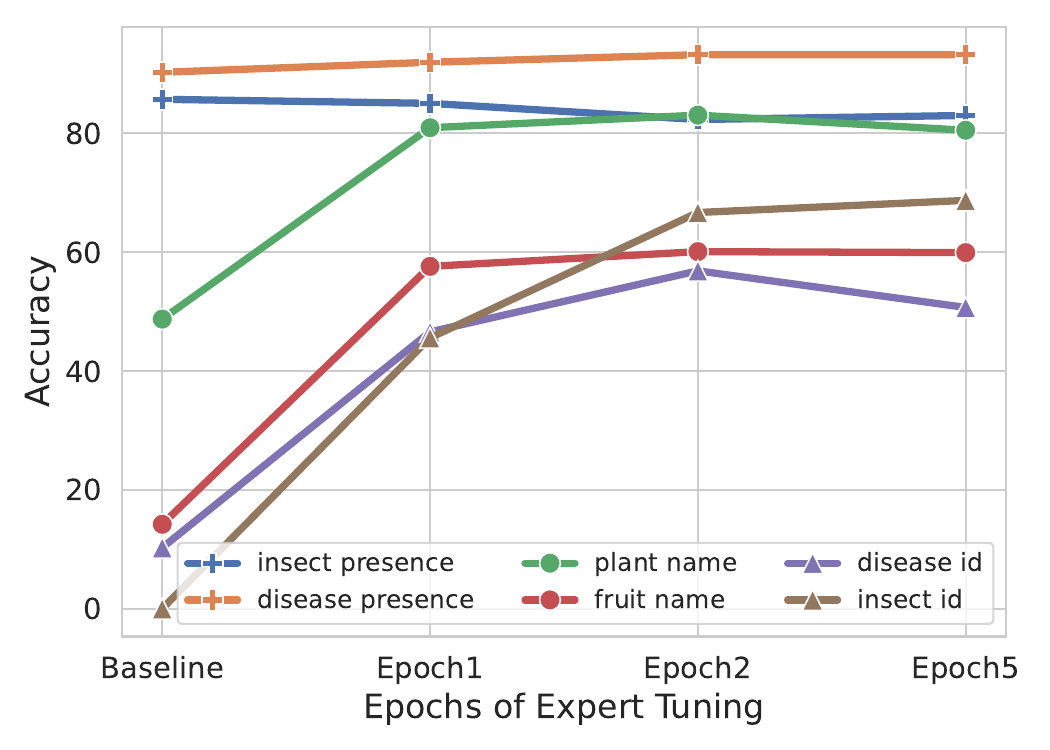}
    \caption{Effect of multiple epochs of expert-tuning on performance on three groups of questions for AgroGPT-3B. }
    \label{fig:epochs}
\end{figure}

\begin{table}
    \centering
    \begin{adjustbox}{width=0.47\textwidth}
    \begin{tabular}{llcc}
    \toprule
    &\textbf{Category} & \textbf{Fully Finetuned} & \textbf{LoRA Finetuned} \\
    \midrule
    \multirow{ 2}{*}{Group1}&Disease Presence & 89.41 & 92.37 \\
    &Insect Presence & 92.52 & 87.75 \\
    \midrule
    \multirow{ 2}{*}{Group2}&Fruits Name & 47.96 & 47.38 \\
    &Plant Name & 72.88 & 72.88 \\

    \midrule
    \multirow{ 2}{*}{Group3}&Disease Id & 51.37 & 43.83 \\
    &Insect Id & 34.01 & 32.65 \\
    \bottomrule
    \end{tabular}
    \end{adjustbox}
     \caption{Comparison of full finetuning and LoRA finetuning for the AgroGPT-7B model across three groups of questions. LoRA performs on par with full finetuning for coarse and general questions from Groups 1 and 2. However, full finetuning performs significantly better than LoRA finetuning for fine-grained and field-specific questions from Group 3.}
     \label{tab:LoRAvsFineTuning}
\end{table}

\noindent \textbf{Expert Evaluations. }
AgroGPT is an expert model that can answer complex agriculture-related questions in detail, making it difficult to assess its performance solely based on quantitative measures. Hence, to further understand AgroGPT's performance, we perform expert evaluations by designing an interface, generating detail-oriented questions from Group 3 (plant diseases and insects that require more fine-grained knowledge of the field), and anonymously presenting answers of baseline and AgroGPT to the expert evaluators. For a thorough description, please refer to the Appendix.
Results are shown in Table~\ref{fig:post_rebuttal}(a), which clearly illustrate that AgroGPT's answers are overwhelmingly preferred as they answer with relevant details. On the other hand, the baseline model often gives generic answers that are less relevant to the problem.}

\noindent \textbf{Expert Tuning after Stage 1 vs. Stage 2}. Another crucial aspect of our method is determining the stage at which expert tuning should be applied.  Our method is based on visual instruction-tuning~\cite{liu2023visual}, which consists of two stages: image-text alignment (stage 1) and general visual instruction tuning (stage 2). We conducted experiments to evaluate the impact of expert tuning after each stage. The results are detailed in Table~\ref{table:stage2vsstage3}. Overall, AgroGPT achieves better results when expert tuning is performed after Stage 2. Further discussion on this topic can be found in the Appendix.

\noindent \textbf{Low-Rank Adaptation vs. Fine-tuning}
Table~\ref{tab:LoRAvsFineTuning} compares Low-Rank Adaptation (LoRA) vs. full fine-tuning for expert tuning. The results show that fine-tuning is significantly better than LoRA fine-tuning. This is in line with previous findings~\cite{biderman2024lora}. 

\subsection{Effect of Number of Expert-tuning Epochs.}
To understand the effect of increasing the number of expert-tuning epochs on the model's performance, we conducted an experiment where we trained AgroGPT-3B for 1, 3, and 5 epochs. Figure~\ref{fig:epochs} compares the results of different numbers of epochs with the baseline (instruction-tuned Mipha3B~\cite{zhu2024mipha}). The results demonstrate that the model's performance on Group 1 questions is not affected by expert tuning. Conversely, more epochs of expert tuning enhance results for Group 2 and Group 3 questions, highlighting AgroInstruct's capability to embed fine-grained and field-specific knowledge into the model. However, the improvement subsides after two epochs, indicating diminishing returns.

\section{Conclusion}
In this work, we introduce AgroGPT, an efficient large vision-language model tailored for the agricultural domain capable of conducting multi-turn, multimodal, open-ended conversations. To achieve this, we present a pipeline for generating expert-tuning data from vision-only datasets, following a self-instruct approach augmented with class-specific external knowledge. Using this pipeline, we constructed the AgroInstruct dataset and trained the efficient conversational model called AgroGPT. We also developed AgroEvals to evaluate the domain-specific capabilities of our model. AgroGPT excels at identifying fine-grained, domain-specific concepts and engaging in detailed conversations on agriculture-specific questions.


{\small
\bibliographystyle{ieee_fullname}
\bibliography{refs}
}

\appendix

\section{Extended Experimental Details}
\subsection{Extended Details of AgroInstruct} For the construction of AgroInstrut, we utilized class-based background knowledge fed to LLM to generate context-based question-answer pairs. A few of the external resources we used for background knowledge are as follows. For farm insects, we took information from \href{https://ag.umass.edu/vegetable/fact-sheets/european-corn-borer}{umass.edu}, \href{https://edis.ifas.ufl.edu/publication/IN790}{ufl.edu}, and \href{https://extension.usu.edu/pests/ipm/notes_ag/veg-western-corn-rootworm}{usu.edu}. For early weed management, we referred to \href{https://www.sare.org/publications/manage-weeds-on-your-farm/velvetleaf/#velvetleaf-management}{sare.org}. For fruits, we consulted \href{https://extension.umn.edu/fruit/raspberry-farming}{umn.edu} and \href{https://anfab.org.au/main.asp?_=Finger%20Lime}{anfab.org.au}.

\begin{figure*}
    \centering
    \includegraphics[width=0.9\linewidth]{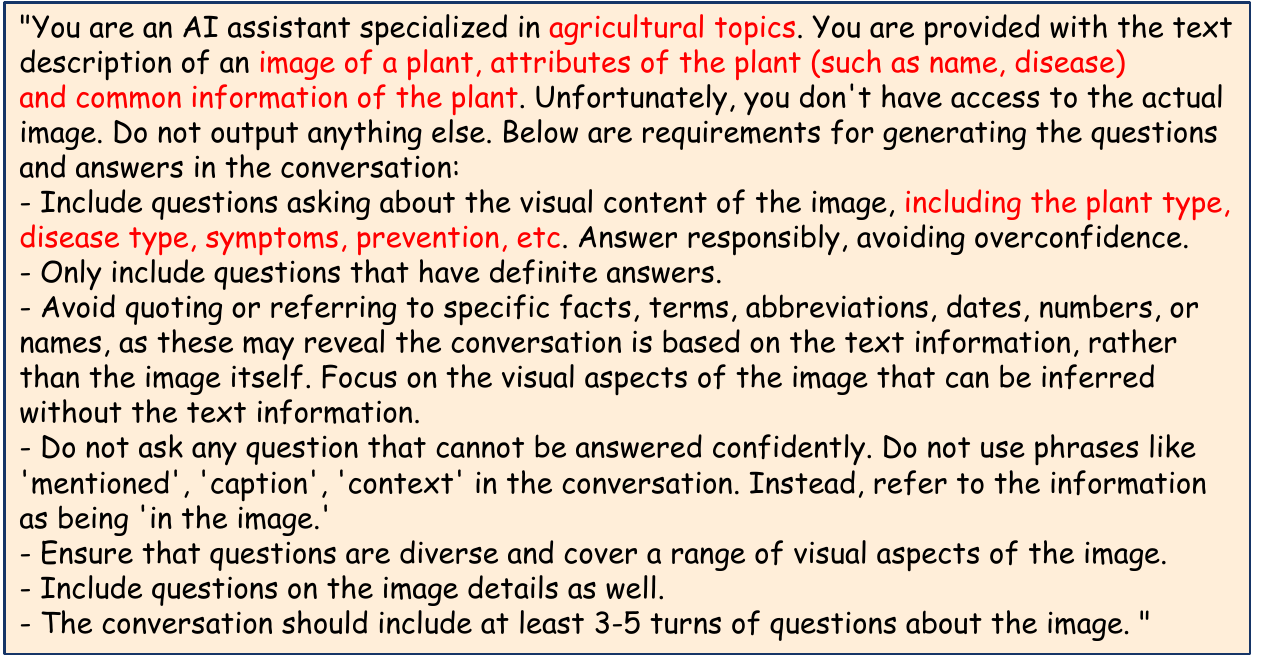} 
    \caption{The system prompt utilized with Mistral~\cite{jiang2023mistral} to generate instruction-tuning (complex questions) data. Mistral is provided with both the system prompt and context to direct its generation of instruction-tuning data. }
    \label{fig:system_prompt}
\end{figure*}

In Figure~\ref{fig:system_prompt}, we have also added the system prompt that we have utilized for generating expert-tuning datasets. Moreover, in Table~\ref{table:AgroEvals}, we briefly describe our different groups of questions from AgroEvals. 

\begin{figure}[h]
    \centering
    \includegraphics[width=0.95\linewidth]{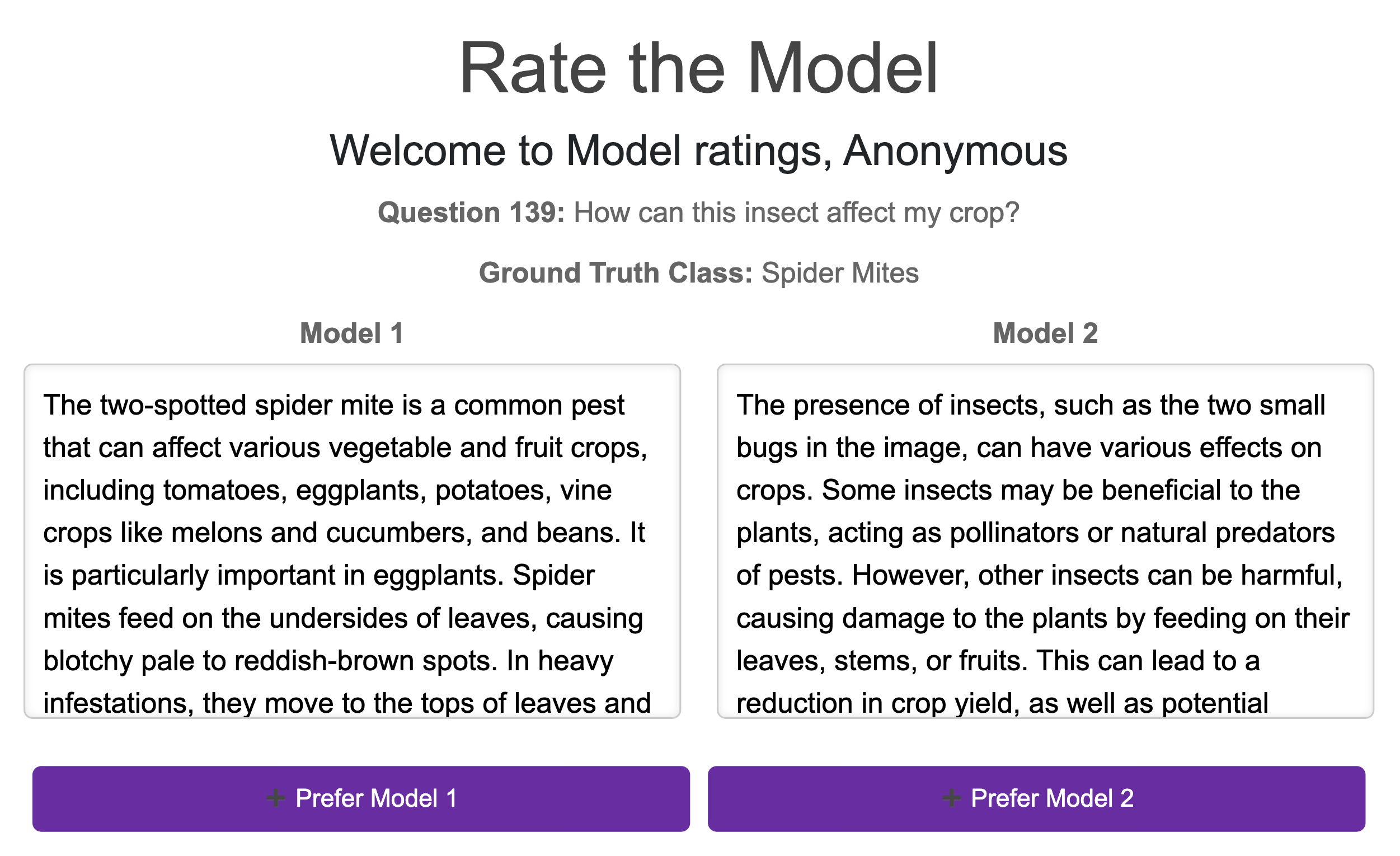}
    \caption{\revision{Our designed interface to get expert evaluations anonymously.}}
    \label{fig:rate_model_app}
\end{figure}

\begin{figure}
    \centering
    \includegraphics[width=0.48\textwidth]{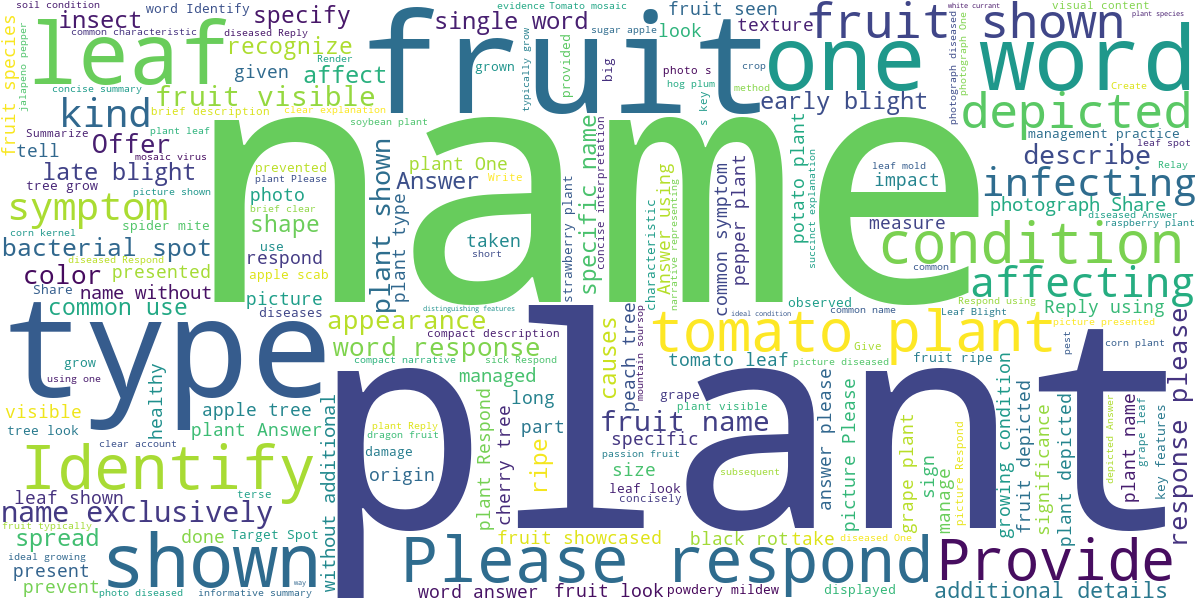}
    (a) Wordcloud for questions in AgroInstruct. 
    \includegraphics[width=0.47\textwidth]{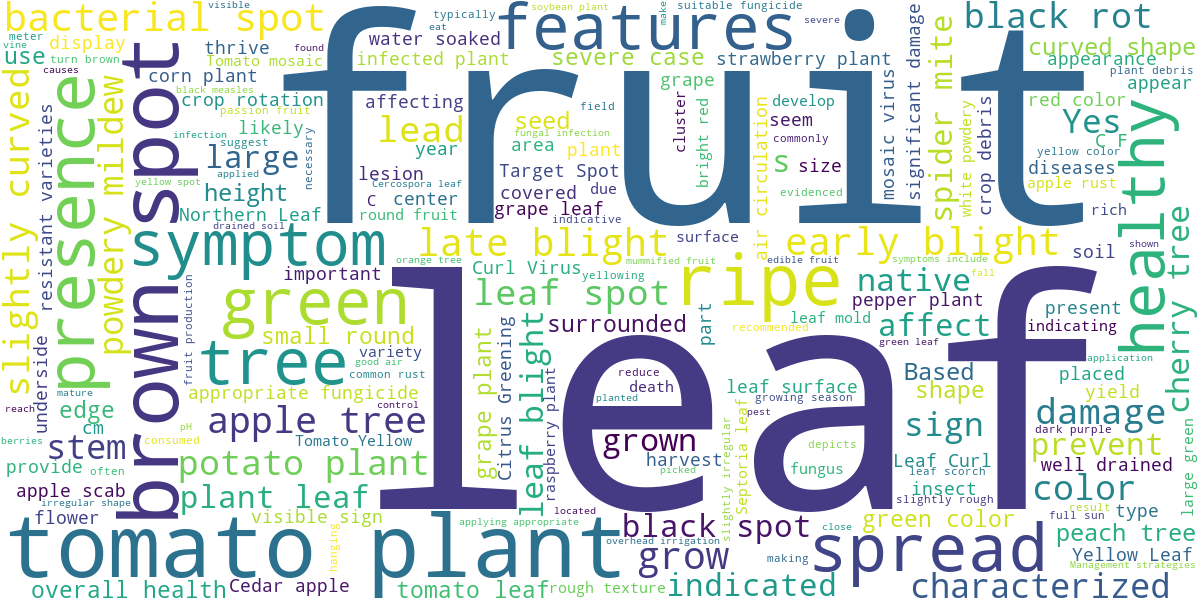}
    (b) Wordcloud for answers in AgroInstruct.
    \caption{A word cloud of constructed expert tuning data with word size based on its frequency in the data.}
    \label{fig:word_cloud}
\end{figure}

\section{Extended Experimental Results}
\label{sec:conversations}

\noindent \textbf{Expert Evaluations. } 
\revision{
AgroGPT is an expert model that can answer complicated and long agriculture-related questions. This makes it difficult to evaluate its performance solely based on quantitative measures. To further understand AgroGPT's performance, we perform expert evaluations, consisting of giving an expert answer to a question generated by both models and getting their preference (e.g., which answer is better). 

To perform this evaluation, we first design a simple interface allowing users to record their preference for different models as shown in Figure~\ref{fig:rate_model_app}. The interface consists of anonymous responses from two models side-by-side. Moreover, it also shows the ground-truth class of the image and the question. This makes it easy for an expert user to see the context and evaluate both models. Moreover, as experts provide feedback on the questions, it automatically takes them to the next question without any input, decreasing friction. 
To evaluate our method's performance on more involved questions, we designed four questions related to diseases and insects since these two topics involve fine-grained information. The four questions are as shown in Table~\ref{tab:questions_expert_evals}. To evaluate these questions, we got help from a person with a background in agriculture inspection. 
As shown in Table~\ref{tab:questions_expert_evals}, AgroGPT's answers are overwhelmingly preferred as it answers with relevant details. The baseline model often gives irrelevant or general answers, while our trained AgroGPT answers with helpful and useful knowledge. }

\begin{table}
    \centering
    \begin{tabular}{ll|p{6cm}}
    \toprule
    Type & No. &  Question \\
    \toprule
         \multirow{3}{*}{Insects} & Q1 & How can this insect affect my crop?  \\
        & Q2 & What are some non-chemical ways to control the infestation of this insect in my field?	\\
         \midrule
       \multirow{3}{*}{Diseases}  &Q1 & What are some biolgoical ways to control this disease?	 \\
         & Q2& How can this disease affect my crop?	\\

         \bottomrule
    \end{tabular}
    \caption{\revision{Four questions utilized for expert evaluations. }}
    \vspace{-10pt}
    \label{tab:questions_expert_evals}
\end{table}

\begin{table}
    \centering
    \resizebox{0.5\textwidth}{!}{
    \begin{tabular}{c|cccc}
    \toprule
        & \multicolumn{2}{c}{Disease} & \multicolumn{2}{c}{Insects}\\
        Baseline &Question 1 &  Question 2 &  Question 1 & Question 2\\
        \midrule
         Baseline & 14\% & 6\% & 6\% & 4\%\\
         AgroGPT & 86\% & 94\% & 94\% & 96\%\\
         \bottomrule
    \end{tabular}}
    \caption{\revision{Results of expert evaluations. Each number represents the percentage of answers given a model that an expert reviewer prefers over others.} }
    \label{tab:expert_evals}
    \vspace{-10pt}
\end{table}

\begin{table}
    \centering
    
    \begin{adjustbox}{width=0.5\textwidth}
    \begin{tabular}{llll}
    \toprule
    &\textbf{Category} & \textbf{Description} & \textbf{Type}\\
    \midrule
    \multirow{ 2}{*}{Group1}&Disease Presence & If a plant is diseased & Coarse \\
    &Insect Presence  & If image has an insect & Coarse \\
    \midrule
    \multirow{ 2}{*}{Group2} & Plant Name & Recognize plant name & Medium\\
            &Fruits Name             & Recognize fruit & Medium\\
    \midrule
    \multirow{ 2}{*}{Group3}
    &Disease Id &  Identify plant disease & Fine-grained\\
    &Insect Id &    Identify insect name & Fine-grained \\
    \bottomrule
    \end{tabular}
    \end{adjustbox}
        \caption{Three different groups of questions, their description, and types of questions used in AgroEvals. }
        \label{table:AgroEvals}
\end{table}

\noindent \textbf{Extended Qualitative Results. }
In the following tables and figures, we present extended qualitative results comparing our model with Bard and LLaVA-34B for conversational capabilities about agriculture. 

\newpage

\begin{table*}[htbp]
  \centering
  \begin{minipage}[b]{0.95\linewidth}
    \centering
     \resizebox{1\textwidth}{!}{
      \begin{tabular}{p{0.1\textwidth}p{0.8\textwidth}}
        \toprule
         & \includegraphics[width=0.2\textwidth]{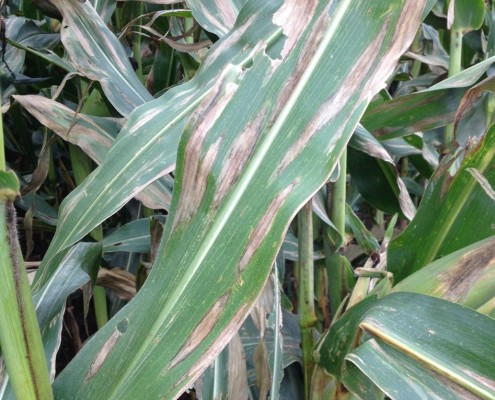}   \\
         \midrule
         User & What is the disease affecting my plant? \\
         AgroGPT & The image shows a corn plant with a yellowish-brown leaf, which is a symptom of a disease called Northern Leaf Blight. \\
         User     & What are the causes of this disease, and what preventive measures can be taken to avoid this disease? \\
         AgroGPT & This disease is caused by the fungus Exserohilum turcicum and can lead to significant yield losses if not managed properly.To prevent this disease, farmers can practice crop rotation, avoid planting susceptible crops in the same field year after year, and use fungicides to control the disease. Additionally, maintaining proper soil moisture and nutrient levels can help reduce the risk of Northern Leaf Blight.\\

         \midrule
         Bard     & I can't give a definitive name but it can be Dickeya:(..) Xanthomonas(..), Fusarium(..), Gibberella(..)  Stem rust: The image also reveals regions with orange pustules, suggesting the presence of stem rust. ...\\
         \midrule
        LLava-34B & The plant in the image appears to be affected by a disease that is causing discoloration and necrosis of the leaves. The symptoms include yellowing and browning of the leaf tissue, which could be indicative of a fungal infection, such as rust or leaf blight, or it could be due to a deficiency in nutrients or water stress. (...) 
    \end{tabular}}
  \end{minipage}
  
  \bigskip
  
  \begin{minipage}[b]{0.95\linewidth}
    \centering
        \resizebox{1\textwidth}{!}{
    \begin{tabular}{p{0.1\textwidth}p{0.9\textwidth}}
    \toprule
         & \includegraphics[width=0.2\textwidth]{figs/armyworm.jpg}   \\
         \midrule
         User & What insect is there in this image? \\
         AgroGPT & The insect in the image is a fall armyworm.\\
         User & How can it affect my crop?? \\
         AgroGPT & Fall armyworms are a type of caterpillar that can cause significant damage to crops, particularly corn and soybeans. They feed on the leaves, stems, and fruits of these plants, which can lead to reduced yield and quality. In severe cases, the infestation can cause complete crop failure. The damage caused by fall armyworms can be difficult to control and can result in significant economic losses for farmers.\\

         \midrule
          Bard    &The insect in the image you sent appears to be a caterpillar, but            it is difficult to identify the exact species without more However, based on  the visible characteristics, here are some possibilities:  Mythimna separata(..), Fall  armyworm (..), Prodenia spp(..), Texas armyworm caterpillar(..) .. \\
         \midrule
        LLava-34B & The insect in the image is a caterpillar.\\
         \bottomrule
    \end{tabular}}
    \caption{Comparison of our efficient AgroGPT  with close-source Bard and 34B LLaVA on open-ended questions about diseased plant and farm insect input images.}
  \end{minipage}
  \label{tab:comapre1}
\end{table*}

\begin{table*}[htbp]
  \centering
  \begin{minipage}[b]{0.95\linewidth}
    \centering
        \resizebox{1\textwidth}{!}{
    \begin{tabular}{p{0.1\textwidth}p{0.9\textwidth}}
    \toprule
         & \includegraphics[width=0.25\textwidth]{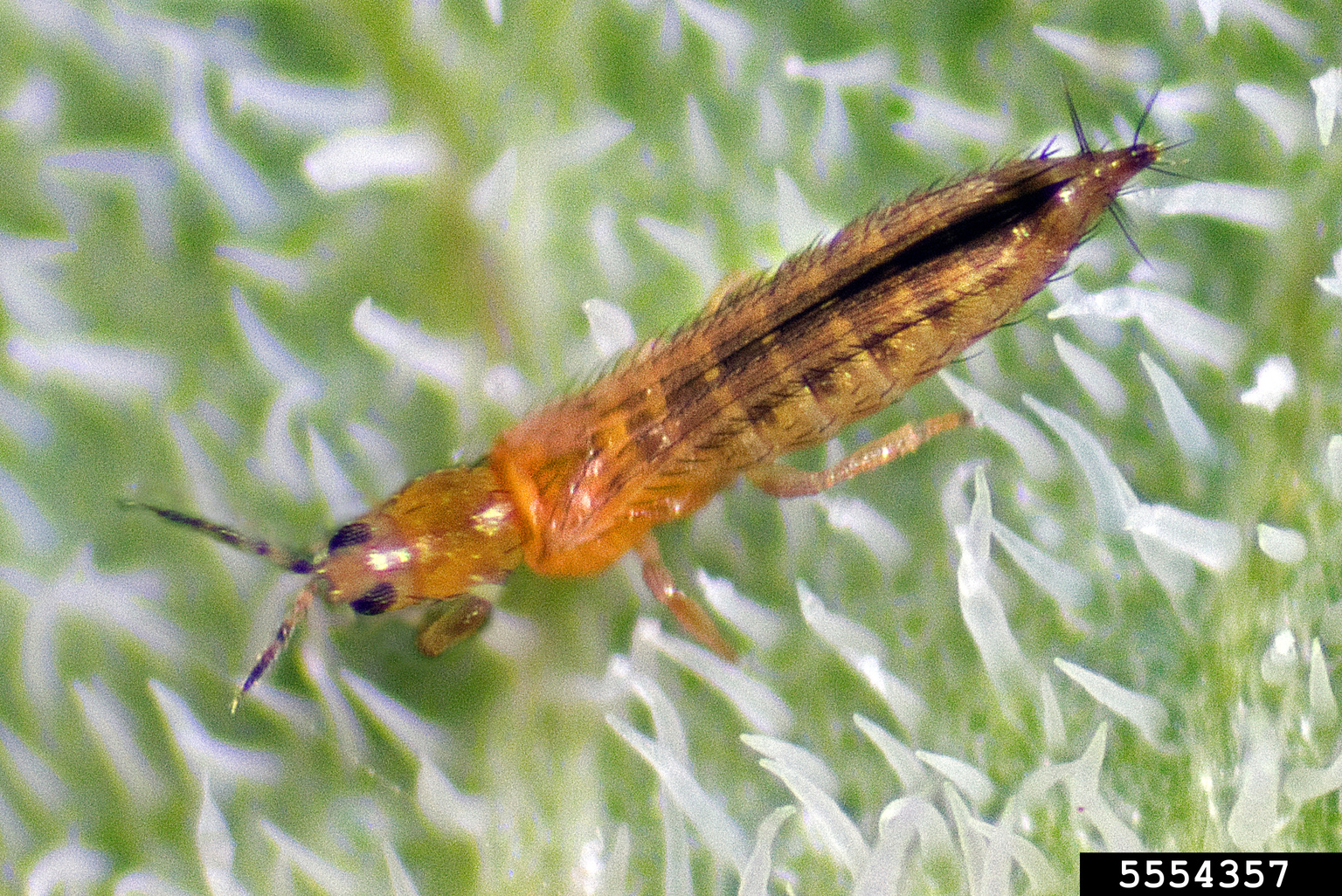}   \\
         \midrule
         User     & What insect is this, and how can it affect my crop? \\
         AgroGPT & The insect in the image is a Thrips, which is a small, slender insect with a long antennae. Thrips are known to feed on the sap of plants, causing damage to the plant tissue. They can 
                  also transmit plant diseases. Thrips can be a significant threat to agriculture, as they can 
                  cause damage to a wide range of crops, including fruits vegetables, and, ornamental plants. The 
                  damage caused by Thrips can lead to reduced crop yields and even crop failure. It is important to 
                  monitor and control Thrips populations to prevent damage to crops and maintain plant health. \\
         User     & What are the biological ways to control these insects? \\
         AgroGPT & There are several biological methods for controlling Thrips, including the use of natural enemies, such 
                  a parasitic wasps, predatory beetles, and predatory mites. These natural enemies can help to reduce 
                   Thrips populations by feeding on their eggs, larvae, and adults. Another method is to use beneficial 
                  nematodes, which can help to control Thrips by parasitizing their eggs and larvae. Additionally, 
                  some plants, such as Thrips-resistant crops, can be used to help reduce Thrips populations.\\

         \midrule
         Bard    & Based on the image you sent, the insect appears to be a caterpillar. This species is a pest of onions, garlic,
                 and other allium crops.  It feeds on the sap of leaves, causing them to become silvered and stunted. \\
         \midrule
       LLava-34B & The insect in the image appears to be a type of caterpillar, possibly a cutworm or armyworm, which are known 
                 to be pests in agriculture. These caterpillars can affect crops in several ways: Feeding Damage(...), 
                  Cutting Plants(...), Transmission of Diseases(...), Damage to Seedlings(...) ... 
                  \end{tabular}
    }
  \end{minipage}
  \label{tab:comapre2}
  \bigskip
  
  \begin{minipage}[b]{0.95\linewidth}
   \centering
  \resizebox{1\textwidth}{!}{
   \begin{tabular}{p{0.1\textwidth}p{0.9\textwidth}}
    \toprule
         & \includegraphics[width=0.25\textwidth]{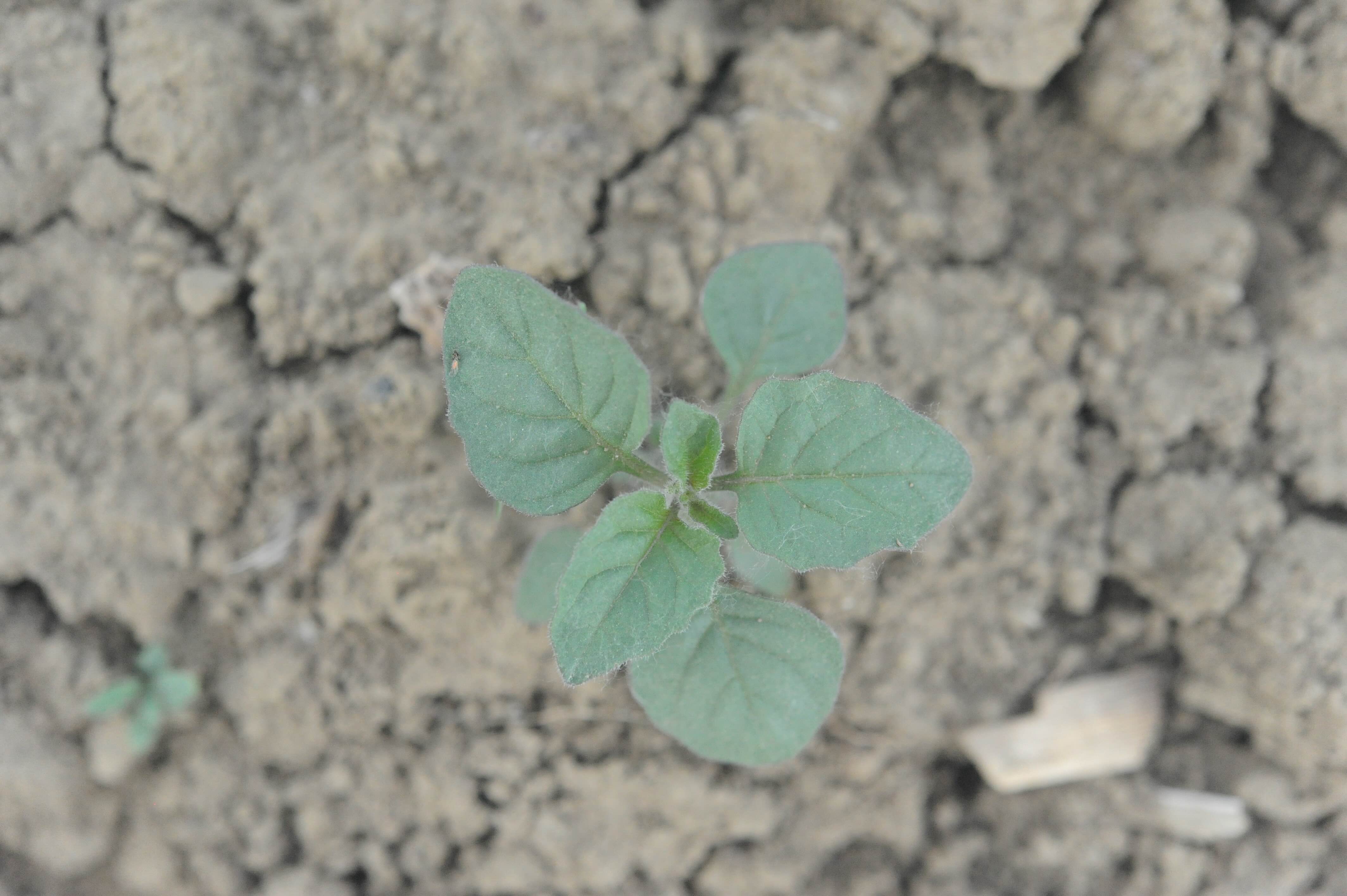}   \\
         \toprule
         \textbf{User}:     & What plant is this? Can this affect the yield of my crop? \\
         \textbf{AgroGPT}:  & The plant in the image is a {\color{blue}young black nightshade}, a weed that can negatively impact crop yields. Black nightshade competes with crops for resources and can reduce crop quality by staining and affecting the taste of produce. It is important to manage black nightshade populations to minimize its impact on agriculture. \\
         \textbf{User}:      & What are some ways to control this weed? \\
         \textbf{AgroGPT}:   & Some ways to control black nightshade include using good cropping practices, such as high seeding rates and narrow row spacing, to create dense, uniform stands that will suppress the weed. No-till systems can also help reduce nightshade densities by minimizing soil disturbance. Herbicides can be used for spot treatments, but care should be taken to avoid harming beneficial plants and to follow label instructions carefully. Solarization, which involves covering the soil with a clear film to block sunlight, can also be effective in killing nightshade seeds. \\
    \midrule
         \textbf{Bard}:      & The plant in the image is most likely a {\color{red}tomato plant (Solanum lycopersicum)}.  Depending on the variety,  tomatoes can (...)
        Whether this tomato plant will affect the yield of your crop depends on several factors, including: (...)
        If you have a large number of tomato plants, they may crowd out your other crops and reduce their yield.
        Finally, the location of the tomato plants (...)
        In general, it is best to err on the side of caution and remove any volunteer tomato plants (...). \\

    \bottomrule
    \end{tabular}
  } 
  \end{minipage}
  \caption{Comparison of our efficient AgroGPT  with close-source Bard and 34B LLaVA on open-ended questions about a farm insect and weed input images on open-ended questions on a holdout image from Farm Insects~\cite{mohan2022insects} and Early Weed Detection~\cite{di2017automatic} datasets. {For weed image, AgroGPT-3B correctly identifies the plant as a weed harmful to crops and suggests effective control measures. In contrast, Bard incorrectly identifies the plant as a tomato and provides irrelevant suggestions.}}
\end{table*}

\begin{table*}
  \centering
  \begin{minipage}[b]{0.95\linewidth}
    \centering
    \resizebox{1\textwidth}{!}{
    \begin{tabular}{p{0.1\textwidth}p{0.8\textwidth}}
    \toprule
         & \includegraphics[width=0.2\textwidth]{figs/tomato_curl_virus.jpg}   \\
         \midrule
         User     & What is the disease affecting the leaves? \\
         AgroGPT & The leaf in the image appears to be affected by Tomato Leaf Curl Virus\\
         User     & How can I control this disease? \\
         AgroGPT & Tomato Leaf Curl Virus is a viral disease that can be controlled \\ 
                  & by planting resistant varieties, removing infected plants and burning them, avoiding \\
                  & the use of infected seedlings, and practicing crop rotation.\\

         \midrule
          Bard    &
                    Based on the image you sent, it is difficult to say for sure what disease is affecting the leaf However, there are a few possibilities, including: Early blight(..) Septoria leaf spot(..), Bacterial leaf spot (..) It is also possible that the leaf is damaged by insects or other pests. \\
         \midrule
        LLava-34B & The plant in the image appears to be affected by a disease that is causing the leaves to become discolored and possibly wilted. The symptoms include yellowing and browning of the leaf edges,which could indicate a variety of issues such as nutrient deficiency, water stress, or a fungal or bacterial infection 
    \end{tabular}}
  \end{minipage}

  \bigskip
  
  \begin{minipage}[b]{0.95\linewidth}
    \centering
    \resizebox{1\textwidth}{!}{
    \begin{tabular}{p{0.1\textwidth}p{0.8\textwidth}}
    \toprule
         & \includegraphics[width=0.2\textwidth]{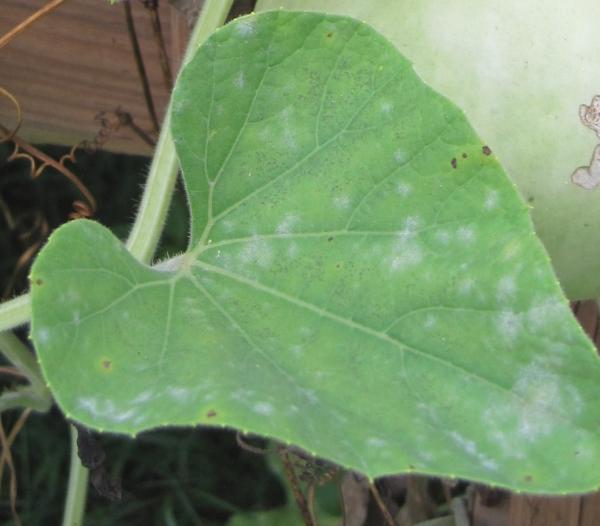}   \\
         \midrule
         User & Is the leaf infected? If yes what is the disease? \\
         AgroGPT & Yes, the leaf in the image is infected with powdery mildew.\\
         User & Is there any way natural way to cure this disease without using artificial pesticides?  \\
         AgroGPT & Yes, there are several natural ways to prevent and treat powdery mildew on squash plants without using artificial pesticides. Planting resistant varieties, Providing good air circulation and drainage,  Avoiding overhead irrigation, Removing infected leaves and debris, 
using organic fertilizers\\

         \midrule
          Bard    & Based on the image you sent, the leaf is likely infected with a fungal  disease called anthracnose. Anthracnose is a common disease that affects many plants, including corn, beans, peppers, and tomatoes. It is caused by a fungus that produces spores that can spread through the air, water, or on infected tools or clothing. \\
         \midrule
        LLava-34B & The leaf in the image appears to be affected by a disease or pest damage. The symptoms include white spots or patches on the leaf surface, which could be indicative of a fungal infection such as powdery mildew or a bacterial infection like bacterial leaf spot. Additionally, there are some brown spots and necrotic tissue, which could be due to a variety of issues, including bacterial or fungal infection, nutrient deficiency, or physical damage. \\
         \bottomrule
    \end{tabular}}

  \end{minipage}
      \caption{Comparison of our efficient AgroGPT with close-source Bard and 34B LLaVA on open-ended questions about two plant input images.}
      \label{tab:compare3}
\end{table*}

\begin{table*}
  \centering
  \begin{minipage}[b]{0.95\linewidth}
    \centering
    \resizebox{1\textwidth}{!}{    \begin{tabular}{p{0.15\textwidth}p{0.8\textwidth}}
    \toprule
         & \includegraphics[width=0.1\textwidth]{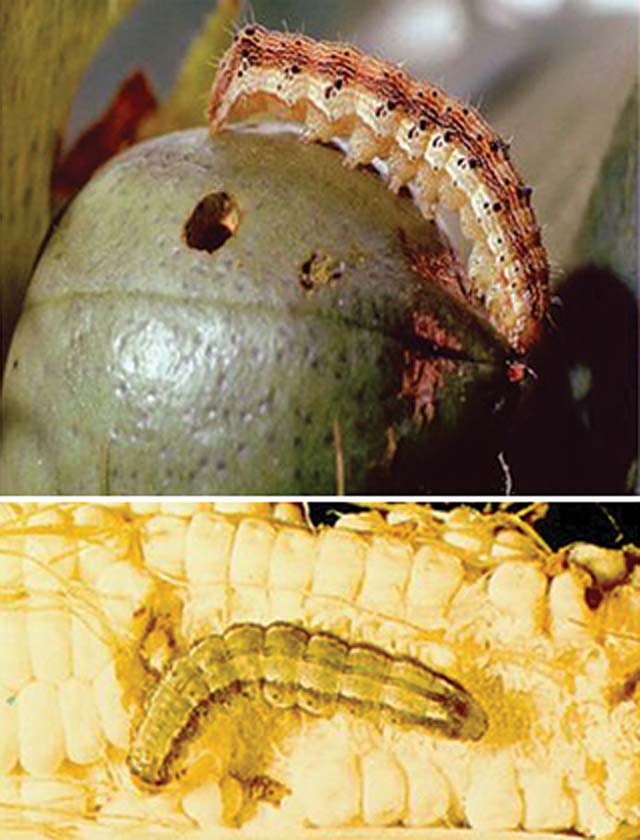}   \\
         \midrule
         User & How can these insect in the image affect the crops? \\
         AgroGPT & Corn earworms are known to feed on the corn plant, causing damage to the crop. They can strip the corn of its leaves, leading to reduced yield and potential crop failure. The larvae of corn earworms can also damage the corn kernels, making them less marketable. Additionally, the presence of these insects can attractother pests, such as birds and rodents, which can further damage the crops.\\
         User & How can these be prevented? \\
         AgroGPT &
        Prevention measures for corn earworms include planting resistant hybrids, using insecticides, and practicing good crop management practices.\\
         \midrule
          Bard    &
                    The insect in the image you sent is most likely a cotton bollworm, also known as Helicoverpa armigera. It's a common pest that affects a wide variety of crops, including cotton, corn, tomatoes, soybeans, and sunflowers. ... \\
         \midrule
        LLava-34B & The insect in the image appears to be a caterpillar,  which is a type of larva of a moth or butterfly. Caterpillars can have a significant impact on crops, particularly when they are in large numbers or when they are feeding on the leaves, stems, or fruits of the plants.
    \end{tabular}
}
  \end{minipage}
  
  \bigskip
  
  \begin{minipage}[b]{0.95\linewidth}
    \centering
    \resizebox{1\textwidth}{!}{    \begin{tabular}{p{0.15\textwidth}p{0.8\textwidth}}
    \toprule
         & \includegraphics[width=0.1\textwidth]{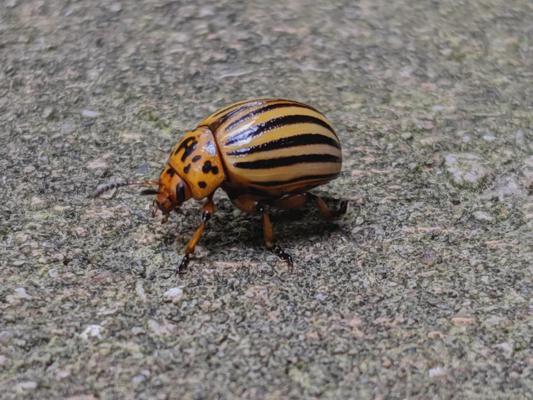}   \\
         \midrule
         User & What disease is affecting this plant?\\
          \includegraphics[width=0.02\linewidth]{figs/icon_3.png} AgroGPT & The plant in the image is affected by early blight, a fungal disease.\\
         User & How can I cure this disease?
\\
          \includegraphics[width=0.02\linewidth]{figs/icon_3.png} AgroGPT& The insect in the image is a Colorado Potato Beetle.\\
           User & How can this insect affect my crop? \\ 
             & Colorado Potato Beetles can cause significant damage to potato\\
             & and eggplant crops. Both adults and larvae feed on the leaves, but larvae damage \\
             &is the most severe. Potatoes can tolerate up to 20\% defoliation, \\
             &while eggplant is more sensitive and can only \\
             &tolerate up to 15\% defoliation before yield is affected. \\
            User& Give me a few ways to control this pest? \\
            AgroGPT & There are several ways to control Colorado Potato Beetles. These include: Cultural practices: Use row covers to protect young plants from beetles. Thinning the crop can also help reduce the number of beetles.Biological control: Lady beetles and lacewings are natural enemies of Colorado Potato Beetles. Encourage their presence by not using insecticides. Chemical control: Selective materials are available for controlling Colorado Potato Beetles.These include pyrethroids, spinosad, and neonicotinoids. It's important to use these materials carefully to minimize harm to beneficial insects. \\
          \midrule
          LLaVA & The insect in the image is a grasshopper.\\
         \midrule
         Bard & Based on the image you sent, the plant is most likely affected by a bacterial disease. There are a few possibilities, including: Bacterial leaf spot: ... Xanthomonas spot: ... Erwinia blight: ...\\
         \midrule
         LLava-34B & The insect in the image is a ladybug, also known as a ladybird. It's a type of beetle known for its distinctive spotted wing covers and its role as a beneficial insect in gardens, where it feeds on aphids and other pests. \\
         \bottomrule
    \end{tabular}}

  \end{minipage}
      \caption{Comparison of our efficient AgroGPT with close-source Bard and 34B LLaVA on open-ended questions about two plant input images.}
      \label{tab:compare3}
\end{table*}

\end{document}